\documentclass[lettersize,journal]{IEEEtran}

\usepackage{amsmath,amsfonts}
\usepackage{algorithmic}
\usepackage{algorithm}
\usepackage{array}
\usepackage[caption=false,font=normalsize,labelfont=sf,textfont=sf]{subfig}
\usepackage{textcomp}
\usepackage{stfloats}
\usepackage{url}
\usepackage{verbatim}
\usepackage{graphicx}
\usepackage{cite}
\usepackage{cuted}    

\newcommand{\methodname}{TAMEn\xspace}
\usepackage[utf8]{inputenc} %
\usepackage[T1]{fontenc}    %
\usepackage{url}            %
\usepackage{booktabs}       %
\usepackage{amsfonts}       %
\usepackage{nicefrac}       %
\usepackage{microtype}      %
\usepackage{xcolor}         %
\usepackage{pifont}
\usepackage{tabularx}
\usepackage{array}

\usepackage{enumitem}
\usepackage{graphicx}
\usepackage{float}
\usepackage{wrapfig}
\usepackage{caption, subcaption, overpic, textpos}
\usepackage{makecell}
\usepackage{comment}
\usepackage{amsmath}
\usepackage{amssymb}
\usepackage{multicol,multirow}
\usepackage{xspace}
\usepackage{lipsum}

\makeatletter
\DeclareRobustCommand\onedot{\futurelet\@let@token\@onedot}
\def\@onedot{\ifx\@let@token.\else.\null\fi\xspace}

\makeatother

\usepackage[svgnames]{xcolor}
\usepackage[table]{xcolor}

\definecolor{green}{RGB}{0,150,10}
\definecolor{blue}{RGB}{0,148,181}
\definecolor{orange}{RGB}{194,153,107}
\definecolor{darkblue}{RGB}{0,50,195}
\definecolor{myblue}{rgb}{0.21,0.49,0.74} 

\usepackage[breaklinks,colorlinks,linkcolor=red,urlcolor=magenta,citecolor=blue]{hyperref}
\usepackage[capitalise]{cleveref}
\crefname{section}{Sec.}{Secs.}
\Crefname{section}{Section}{Sections}
\Crefname{figure}{Figure}{Figures}
\crefname{figure}{Fig.}{Figs.}
\Crefname{table}{Table}{Tables}
\crefname{table}{Tab.}{Tabs.}

\begin{document}

\title{\methodname: Tactile-Aware Manipulation Engine\\ for Closed-Loop Data Collection in Contact-Rich Tasks}
\author{
    \textbf{Longyan Wu} $^{1,2,3}$ \quad
    \textbf{Jieji Ren} $^{4}$ \quad
    \textbf{Chenghang Jiang} $^{5}$ \quad
    \textbf{Junxi Zhou} $^{5}$ \\
    \textbf{Shijia Peng} $^{3}$ \quad 
    \textbf{Ran Huang} $^{1}$ \quad
    \textbf{Guoying Gu} $^{4}$ \quad
    \textbf{Li Chen} $^{3}$ \quad
    \textbf{Hongyang Li} $^{2,3}$ \\
    $^{1}$ Fudan University\quad
    $^{2}$ Shanghai Innovation Institute\quad
    $^{3}$ OpenDriveLab at The University of Hong Kong \\ 
    $^{4}$ Shanghai Jiao Tong University\quad
    $^{5}$ East China University of Science and Technology\quad
    \url{https://opendrivelab.com/TAMEn}
}

\maketitle

\begin{strip}
\centering
  \begin{minipage}{\textwidth}
    \centering
    \includegraphics[width=\textwidth]{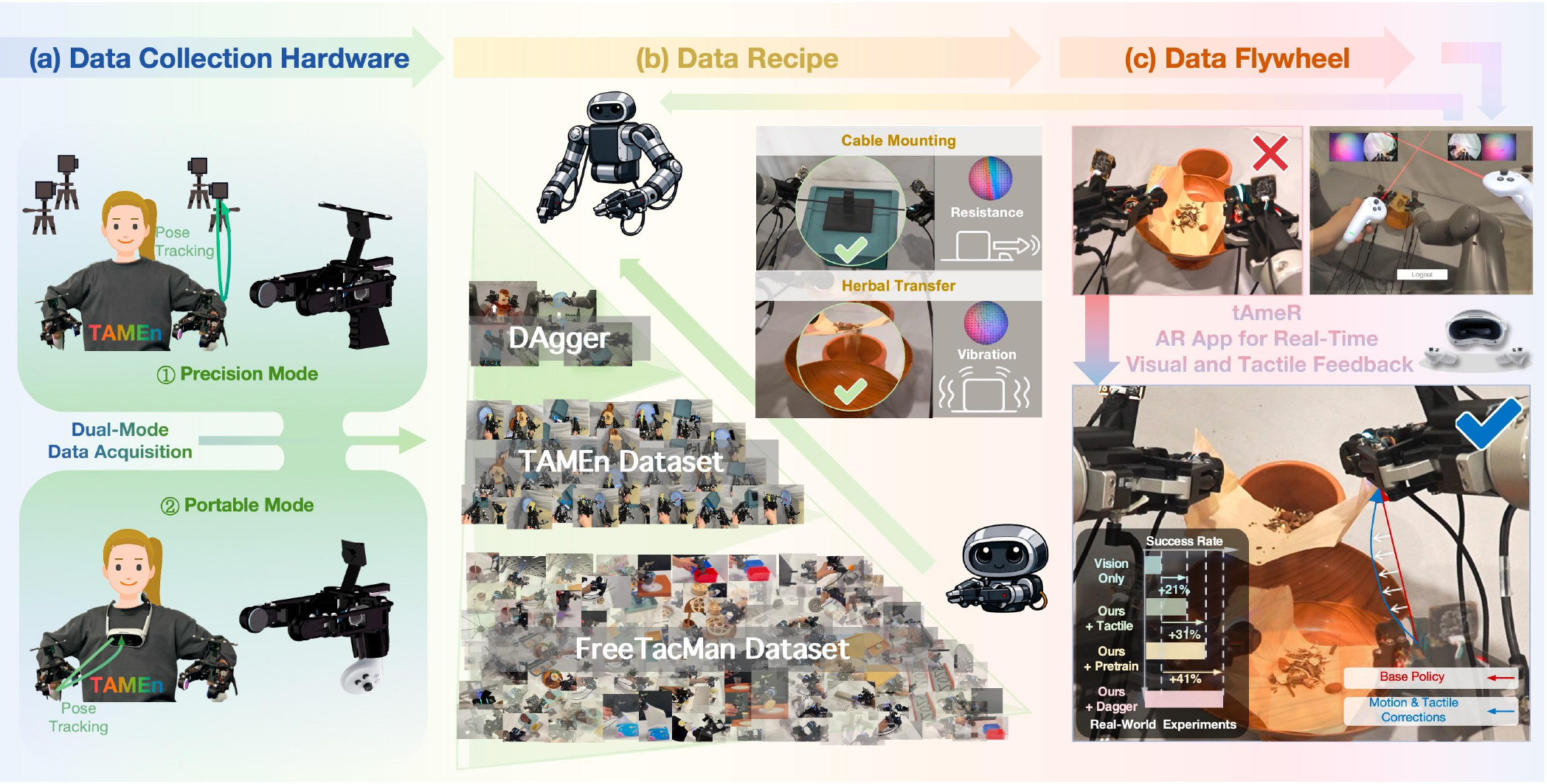}%
    \captionof{figure}{%
     Introducing \textbf{\methodname}, a \textbf{T}actile-\textbf{A}ware \textbf{M}anipulation \textbf{En}gine for closed-loop data collection in contact-rich bimanual tasks, which builds upon the UMI paradigm with key enhancements in multimodality, precision-portability synergy, replayability, and data flywheel.
(a) Wearable visuo-tactile interface captures rich multimodal data while breaking the precision-portability trade-off through a dual-mode pipeline that fast switches between MoCap and VR-based tracking.
(b) Online feasibility checking ensures demonstrations are reliably replayable on robot. All data are unified into a pyramid for efficient staged learning across generalization, coordination, and failure recovery.
(c) \textbf{tAmeR}, our AR-based teleoperation system, helps collect recovery data with tactile feedback during policy execution and feeds them back into the pyramid for continuous policy refinement.
    }\label{fig:teaser}
  \end{minipage}%
\end{strip}

\begin{abstract}
Handheld paradigms offer an efficient and intuitive way for collecting large-scale demonstrations of robot manipulation. However, achieving contact-rich bimanual manipulation through these methods remains a pivotal challenge, which is substantially hindered by hardware adaptability and data efficacy. 
Prior hardware designs remain gripper-specific and often face a trade-off between tracking precision and portability. Furthermore, the lack of online feasibility checking during demonstration leads to poor replayability.
More importantly, existing handheld setups struggle to collect interactive recovery data during robot execution, lacking the authentic tactile information necessary for robust policy refinement.
To bridge these gaps, we present~\methodname, a tactile-aware manipulation engine for closed-loop data collection in contact-rich tasks. 
Our system features a wearable interface that enables rapid adaptation across heterogeneous grippers. To balance data quality and environmental diversity, we implement a dual-mode acquisition pipeline: a precision mode leveraging motion capture for high-fidelity demonstrations, and a portable mode utilizing VR-based tracking for in-the-wild acquisition and tactile-visualized recovery teleoperation. Building on this hardware, we unify large-scale tactile pretraining, task-specific bimanual demonstrations, and human-in-the-loop recovery data into a pyramid-structured data regime, enabling closed-loop policy refinement.
Experiments show that our feasibility-aware pipeline significantly improves demonstration replayability, and that the proposed visuo-tactile learning framework increases the average task success rate from 34\% to 75\% across diverse bimanual manipulation tasks.
We further open-source the hardware and dataset to facilitate reproducibility and support research in visuo-tactile manipulation.

\end{abstract}


\section{Introduction}
\IEEEPARstart{P}{hysical} interaction is fundamental to contact-rich manipulation, especially in bimanual tasks where two end-effectors must coordinate through shared objects under changing force~\cite{li2025vitaminbreliableefficientvisuotactile}, deformation~\cite{hu2023occlusion}, and support conditions~\cite{Kim_2024}. 
In these scenarios, success often depends on subtle contact events, such as contact onset, excessive loading, and incipient slip~\cite{10598389}, which are difficult to infer from vision alone~\cite{zhai2026skillvla,yang2026riseselfimprovingrobotpolicy,intelligence2025pi05visionlanguageactionmodelopenworld}. 
Tactile sensing therefore plays a critical role in enabling robust manipulation~\cite{11024242,feng2026anytouch2generaloptical}. However, unlike visual data, which can be collected at scale from internet videos or human recordings~\cite{11127989,hoque2026egodexlearningdexterousmanipulation,shi2026egohumanoidunlockinginthewildlocomanipulation}, tactile data must be generated through direct physical interaction. These challenges motivate the development of hardware interfaces for efficient, high-quality, and scalable visuo-tactile data collection.

Existing data collection pipelines remain limited in both interaction fidelity and hardware adaptability. Most teleoperation systems rely primarily on visual feedback and thus fail to provide operators with continuous, fine-grained tactile information~\cite{xue2025reactivediffusionpolicyslowfast,li2026unibidexunifiedteleoperationframework,11128299}. 
This limitation is especially pronounced in bimanual manipulation, where operators often struggle to judge contact states and have to rely on repeated visual confirmation and corrective adjustments, reducing data collection efficiency.
Handheld paradigms enable efficient and intuitive data collection~\cite{chi2024universal,ha2024umilegsmakingmanipulation,rayyan2025mvumiscalablemultiviewinterface,yu2026egomilearningactivevision,zhaxizhuoma2025fastumiscalablehardwareindependentuniversal,10855557,huang2025umigenunifiedframeworkegocentric}, and recent advances have integrated tactile sensors to record contact information~\cite{helmut2025tactileconditioneddiffusionpolicyforceaware,11124252,lee2025manipforceforceguidedpolicylearning,li2026simultaneoustactilevisualperceptionlearning}.
However, many existing designs are tailored to a specific end-effector design, making them difficult to transfer to grippers with different kinematic structures and geometric parameters.
This calls for a systematic design abstraction that elevates demonstration-interface design from instance-specific adaptation to configuration-level mapping~\cite{wu2026dexgraspzeromorphologyalignedpolicyzeroshot,lee2026xgraspgripperawaregraspdetection}.

   \begin{figure*}[!t]
      \centering
      \includegraphics[width=\linewidth]{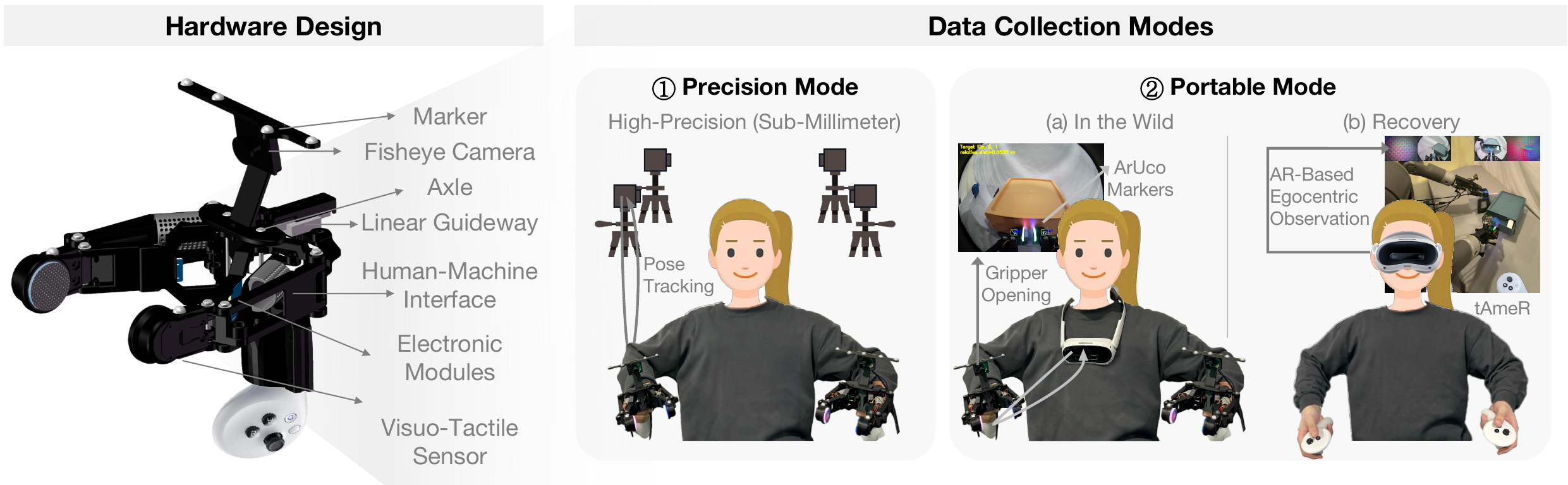}
      \caption{\textbf{Hardware system. } \textit{Left}: Structure of~\methodname. \textit{Right}: two data collection modes, supporting high-precision demonstration collection and portable in-the-wild or recovery data acquisition.
 }
      \label{fig:datacollect}
   \end{figure*}

Successful handheld demonstrations do not necessarily translate into executable robot behavior~\cite{chen2024arcap,fang2026robopocket}. Especially in bimanual manipulation, collected trajectories are more likely to violate inverse kinematics, joint-limit, workspace, or motion constraints due to increased motion complexity~\cite{11339196}.
In such cases, a demonstration may appear successful at the collection stage but fail to replay on robot, leading to substantial offline filtering and manual data cleaning. 
These limitations motivate data acquisition pipelines that integrate robot-side feasibility into collection rather than leaving it to offline post-processing.
Beyond executability, successful trajectories alone are still insufficient for handling near-failure states~\cite{xu2025compliantresidualdaggerimproving,dai2025racer}.
In bimanual contact-rich manipulation, 
near-failure behavior is marked by subtle contact changes such as force buildup, incipient slip, local deformation, or unstable object support~\cite{hu2025racrobotlearninglonghorizon}. Such signals emerge only through real physical interaction and are difficult to reproduce faithfully through offline demonstrations or observe-then-collect corrections~\cite{han2026dexhil}. 
This motivates a data collection pipeline that complements nominal demonstrations with recovery data near realistic failure states.

In this work, we introduce \textbf{\methodname}, a visuo-tactile manipulation data engine for closed-loop data collection in bimanual contact-rich tasks. It enables efficient visuo-tactile bimanual data collection through a human-machine interface. Its dual-mode acquisition pipeline ensures high-quality motion capture for precision tasks while supporting scalable in-the-wild data collection. By incorporating real-time feasibility validation, the system ensures that collected demonstrations are reliably replayable on robot, eliminating the need for costly offline filtering. Beyond nominal demonstrations, it also enables recovery-oriented data collection during robot execution with authentic tactile feedback via AR-based teleoperation (tAmeR), enabling continuous policy refinement. To efficiently leverage these heterogeneous data sources, we introduce a pyramid-structured data regime that provides broad tactile priors from large-scale single-arm data, coordination-aware behaviors from task-specific bimanual demonstrations, and robust recovery capabilities from realistic failure states.

In summary, our main \textbf{contributions} are: 

\textbf{(i)}
A visuo-tactile data \textit{engine} for bimanual contact-rich manipulation, which integrates hardware, acquisition strategy, and policy learning into a closed-loop framework.

\textbf{(ii)}
A human-machine \textit{interface} that supports a dual-mode pipeline with sub-millimeter MoCap and VR-based in-the-wild acquisition, and can rapidly adapt to heterogeneous grippers.

\textbf{(iii)}
A data collection \textit{recipe} that incorporates real-time validation during collection and organizes heterogeneous multimodal data into a pyramid-structured regime for staged learning.

\textbf{(iv)}
A closed-loop data \textit{flywheel} that leverages AR-based teleoperation with tactile feedback (tAmeR) to refine policies using corrective data from realistic failures.

\IEEEpubidadjcol
\section{Related Work}

\subsection{Data Collection Interfaces for Robotic Manipulation}
Collecting large-scale high-quality demonstrations for contact-rich manipulation remains challenging.
While teleoperation offers precise control
and direct mapping to robot embodiments, its reliance on a physical robot makes data collection costly and difficult to scale across unstructured environments~\cite{bu2025agibot,kwon2025humanoidvisualtactileactiondatasetcontactrich,fu2024mobilealohalearningbimanual,chen2025lemmoplanllmenhancedlearningmultimodal}. It is also inefficient for fine-grained manipulation, where precise alignment may require repeated back-and-forth adjustments~\cite{han2026dexhil}. This trial-and-error process may introduce ambiguous supervision for policy learning and increase the risk of hardware damage. 
Handheld data collection interfaces, such as UMI-style systems~\cite{chi2024universal,ha2024umilegsmakingmanipulation,rayyan2025mvumiscalablemultiviewinterface,yu2026egomilearningactivevision,zhaxizhuoma2025fastumiscalablehardwareindependentuniversal,10855557,huang2025umigenunifiedframeworkegocentric}, offer a more natural and efficient alternative for acquiring large-scale datasets~\cite{generalist2025gen0,wu2025freetacman,liu2025fastumi100kadvancingdatadrivenrobotic,zhu2025touchwildlearningfinegrained}. Recent handheld interfaces have also begun to integrate tactile sensing to capture fine-grained contact information during manipulation~\cite{helmut2025tactileconditioneddiffusionpolicyforceaware,11124252,lee2025manipforceforceguidedpolicylearning,li2026simultaneoustactilevisualperceptionlearning,choi2026wild}.
Visuo-tactile sensors are particularly appealing in this context, as they provide high-resolution, sensitive multimodal observations while remaining easy to integrate into policy learning pipelines~\cite{cheng2026tacumi,li2025vitaminbreliableefficientvisuotactile,ren2023mc,zheng2026omnivtavisuotactileworldmodeling}.
Despite this progress, existing handheld systems still face practical challenges in tracking, executability, and cross-morphology deployment.
There is a trade-off between accuracy and portability. High-precision systems, such as optical motion capture~\cite{wu2025freetacman} and vive trackers~\cite{zhang2026touchguide}, depend on external base stations and are therefore less suitable for in-the-wild data collection. More portable alternatives, including SLAM-based methods~\cite{zhu2025touchwildlearningfinegrained,liu2025vitaminlearningcontactrichtasks,fang2026robopocket} and VR tracking~\cite{li2025vitaminbreliableefficientvisuotactile,xu2025exumi}, often suffer from scene dependence or insufficient precision for contact-rich manipulation.
In addition, demonstrations collected through open-loop handheld recording are not necessarily executable on the target robot and therefore often require replay-based validation~\cite{fang2026robopocket}. Moreover, existing handheld collectors are typically tailored to a particular gripper design, requiring users either to use the same end-effector or to redesign the collector for a different gripper.

\subsection{Human-in-the-Loop Policy Correction and Recovery}
Successful demonstrations alone often provide limited coverage of failure states encountered during execution.
This motivates collecting corrective data to improve policy robustness under covariate shift~\cite{huang2026forceawareresidualdaggertrajectory,ross2011reduction}. Prior methods typically collect such corrective data either by recording offline demonstrations that cover the policy’s typical failure scenarios~\cite{chi2024diffusionpolicyvisuomotorpolicy} or by introducing online human correction during policy execution~\cite{spencer2020learning,yu2026chi0resourceawarerobustmanipulation}.
Teleoperated systems~\cite{wu2025robocopilothumanintheloopinteractiveimitation,chen2025conrftreinforcedfinetuningmethod} enable more seamless intervention. Compliant Residual DAgger~\cite{xu2025compliantresidualdaggerimproving} further improves continuity through compliant on-policy correction. 
More recently, RoboPocket~\cite{fang2026robopocket} enables robot-free corrective data collection through handheld AR-based policy visualization.
However, existing methods either place less emphasis on tactile-centered recovery data or involve physically guiding the robot arm during execution, which can be cumbersome in practice, as moving even a single robot arm may require both hands.
\section{Method}

\subsection{System Overview}
To support closed-loop visuo-tactile learning in contact-rich bimanual manipulation, the system must fulfill three distinct functions: enabling efficient multimodal data collection that accommodates precision and portability, ensuring that collected data are executable on the robot and organized for staged learning, and supporting closed-loop policy refinement through AR-based teleoperation with tactile feedback in realistic failure states.
For data acquisition, the system supports two collection modes: a precision mode for high-fidelity demonstration capture and a portable mode for in-the-wild data collection and recovery. 
During collection, the operator interacts with the environment using the handheld interface while the system records synchronized visual, tactile, and motion data. The tracked motions are checked online for robot executability, so infeasible demonstrations can be identified during collection rather than removed afterward. 
To support recovery during policy execution, we develop tAmeR, an AR app that provides the operator with real-time visual and tactile feedback and enables recovery-oriented teleoperation. 
The resulting data are organized into different levels for representation pretraining, task-specific bimanual learning, and recovery-oriented refinement.

\subsection{Hardware Design}
\label{sec:method-hardware}

\textbf{Modular interface design.}
The proposed interface adopts a shared visuo-tactile hardware backbone that enables modular extension across sensing and tracking components. It is built around a wearable in-situ gripper design, where the operator’s thumb and index finger are coupled to an ergonomically designed rigid–soft structure, allowing natural finger motion during manipulation. An inverted crank-slider mechanism converts the finger-driven motion into synchronized gripper actuation while preserving direct in-situ contact at the fingertip.
On top of this backbone, the system supports modular fingertip sensing modules, allowing a wide range of visuo-tactile sensors to be integrated with only minor local modifications. As shown in~\cref{fig:appendix_multisensor}, we instantiate this design with multiple representative sensors, including GelSight, Xense, DW-Tac, PaXini, and ours. In this work, we use our sensor for validation, which offers modular design, ease of deployment across different robots, robust tactile sensing, and simplified fabrication and calibration.
The same backbone also accommodates interchangeable tracking attachments, enabling rapid switching between motion-capture-based tracking and portable VR-based operation.

\textbf{Dual-mode acquisition configuration.}
The system supports two acquisition modes: a motion-capture mode for high-precision tracking and a portable mode for low-cost deployment in unstructured environments. 
In the motion-capture mode, the interface is tracked by the NOKOV system, enabling sub-millimeter pose tracking. The markers are arranged in a structured layout, among which four markers are mounted above the camera module to improve visibility, while two markers attached to the gripper are used to track the gripper opening distance.
In the portable mode, the marker assembly is replaced by a quick-detachable VR handle mounted on the gripping region, allowing the same interface to be rapidly deployed for in-the-wild data collection. Beyond collection, the detachable handle also supports immediate transition to recovery-oriented robot teleoperation. This makes the portable setup a practical low-cost solution for rapid deployment to new tasks, with a total hardware cost of approximately \$700+ for the dual-arm system.
   \begin{figure*}[!t]
      \centering
      \includegraphics[width=\linewidth]{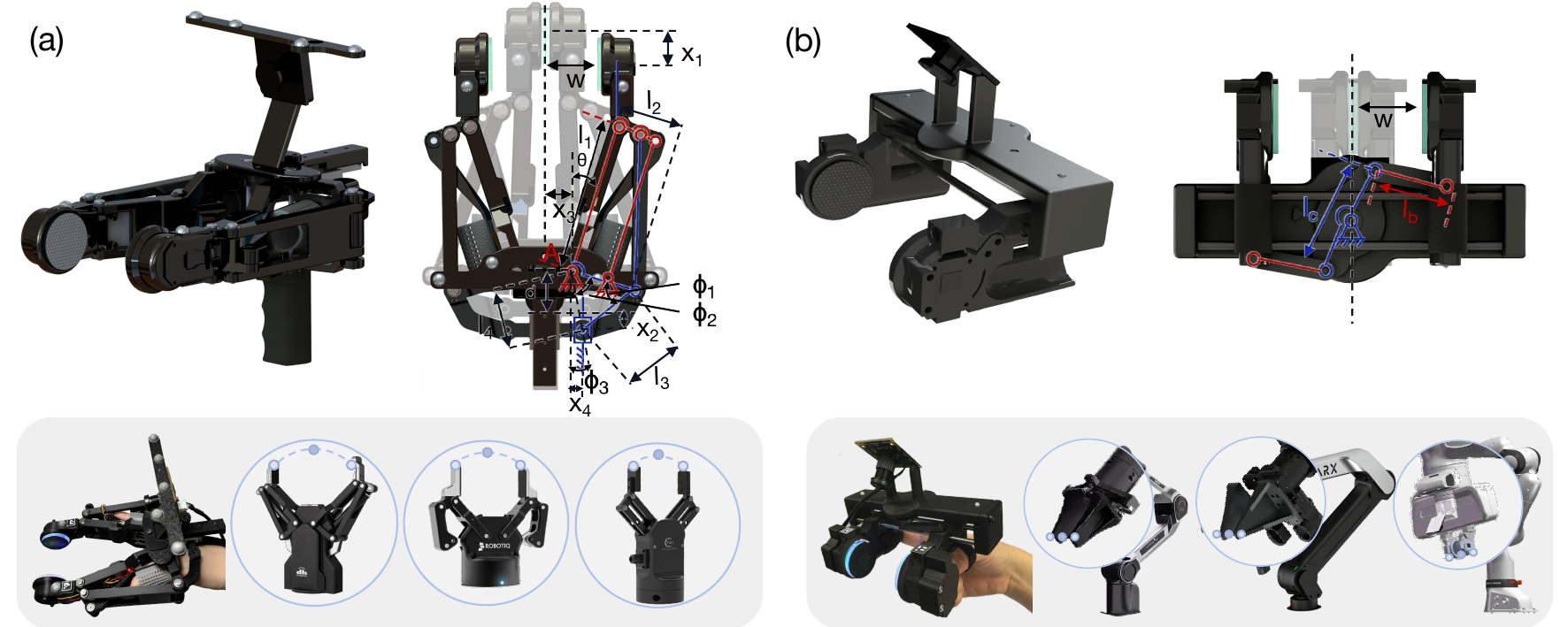}
      \caption{\textbf{Mechanisms of the proposed handheld gripper interface. }(a) Flexion–extension gripper. (b) Parallel-jaw gripper. \textit{Left}: Overall view of the interface. \textit{Right}: Kinematic schematic with key geometric parameters.
 }
      \label{fig:mechanism}
   \end{figure*}
   
\textbf{Interface adaptation across gripper morphologies.}
This design turns handheld collector adaptation into a standardized configuration-level process, reducing the need for gripper-specific redesign and enabling faster deployment across heterogeneous end-effectors. It extracts the key geometric and kinematic characteristics of the target gripper and maps them to the corresponding handheld interface through a unified mechanism template. Instead of requiring a gripper-specific collector or relying on post hoc compensation for morphology-dependent motion differences~\cite{zhaxizhuoma2025fastumiscalablehardwareindependentuniversal}, the interface can be adapted to different gripper structures by adjusting only a small set of geometric parameters. This substantially simplifies the deployment of UMI-style interfaces across heterogeneous robot grippers.~\Cref{fig:mechanism} illustrates the design principle for representative gripper morphologies, including flexion–extension and parallel-jaw grippers.

\textbf{Flexion-extension gripper.}
For the flexion-extension gripper, we characterize its grasping behavior using two key motion quantities: the jaw opening width \(w\) and the fingertip fore-aft displacement \(x_1\) during closure. 
Here, $l_1$, $l_2$, and $l_3$ denote the fixed lengths of the linkages. When the slider moves to its foremost position, its distance to the fixed point $A$ is $d$. The slider displacement from this foremost position is denoted by $x_2$. Accordingly, the instantaneous distance between the slider and point $A$ along the sliding direction is $d+x_2$, and the corresponding Euclidean distance is denoted by $l_4$. In addition, $x_3$ denotes the fixed offset from the slider mounting axis to the gripper symmetry axis, and $x_4$ denotes the distance from the slider mounting axis to the axis passing through point $A$ and parallel to the gripper symmetry axis. The detailed geometric annotations are shown in~\Cref{fig:mechanism}(a).
By vector analysis, the mechanism satisfies:

\begin{equation}
\left\{
\begin{aligned}
w &= x_3 + l_1 \sin\theta \\
x_1 &= l_1 - l_1 \cos\theta \\
l_4 &= \sqrt{x_4^2 + (d + x_2)^2} \\
\phi_3 &= \arctan\left(\frac{x_4}{d + x_2}\right) \\
\phi_2 &= \arccos\left(\frac{l_2^2 + l_4^2 - l_3^2}{2 l_2 l_4}\right) \\
\theta &= \frac{\pi}{2} - \phi_3 - \phi_2
\end{aligned}
\right.
\end{equation}

By jointly solving the above equations, $w$ can be expressed as a function of $x_2$ and $x_3$, and $x_1$ as a function of $x_2$:
\begin{equation}
\begin{aligned}
w(x_2,x_3) = x_3 + l_1 \sin \Bigg[
\frac{\pi}{2}
- \arctan\left(\frac{x_4}{d + x_2}\right) \\
- \arccos\left(
\frac{l_2^2 + x_4^2 + (d + x_2)^2 - l_3^2}
{2 l_2 \sqrt{x_4^2 + (d + x_2)^2}}
\right)
\Bigg]
\end{aligned}
\end{equation}

\begin{equation}
\begin{aligned}
x_1(x_2) =
l_1 \Bigg[
1 - \cos \Bigg(
\frac{\pi}{2}
- \arctan\left(\frac{x_4}{d + x_2}\right) \\
- \arccos\left(
\frac{l_2^2 + x_4^2 + (d + x_2)^2 - l_3^2}
{2 l_2 \sqrt{x_4^2 + (d + x_2)^2}}
\right)
\Bigg)
\Bigg]
\end{aligned}
\end{equation}
Therefore, the mechanism admits a decoupled parameterization for interface adaptation. Given the target maximum fingertip fore-aft displacement $x_1^{\max}$ and jaw opening width $w^{\max}$, we first determine the required slider stroke $x_2^{\max}$ from $x_1^{\max}$, and then choose $x_3$ to satisfy $w^{\max}$. This reduces interface adaptation to specifying only two target motion requirements. The handheld interface used in our experiments follows the flexion–extension configuration.

\textbf{Parallel-jaw gripper.}
A crank–slider mechanism is commonly used in collection interfaces for this gripper type, as illustrated in~\cref{fig:mechanism}(b). Since the motion of the parallel-jaw gripper is dominated by symmetric linear opening and closing, interface adaptation mainly reduces to matching the maximum jaw opening range of the target gripper. This quantity is determined by the crank-slider geometry, with:
\begin{equation}
w_{\max} = l_c + 2l_b,
\end{equation}
where \(l_c\) denotes the crank length and \(l_b\) the driving linkage length. To avoid increasing the overall width of the handheld interface, we fix \(l_c\) and adapt the mechanism to different parallel-jaw grippers by adjusting only \(l_b\).
This provides a simple parameterization for adapting the interface to different parallel-jaw grippers.

\subsection{Feasibility-Aware Data Acquisition Pipeline}
\label{sec:method-datacollect}
\textbf{Structured motion capture tracking.} Reliable motion capture is critical for feasibility-aware data collection. In bimanual contact-rich manipulation, marker occlusion frequently arises from hand-object interaction, inter-arm interference, and environmental clutter, which makes direct marker tracking unstable. To address this issue, we represent each handheld interface as a structured marker object with predefined geometric topology and marker identities. This formulation allows the tracker to exploit structural consistency for pose estimation and marker recovery under partial occlusion or noisy observations.
For each interface configuration, we initialize the structured object model from a short recorded sequence of unlabeled markers. Marker identities and their structural connectivity are established in post-processing, with the first frame used to determine marker correspondence. Based on this initialization, correction-based tracking is applied to propagate marker identities throughout the sequence, while local repair is performed only on segments affected by occlusion or identity ambiguity. Once the sequence is fully labeled, the object model is then constructed, enabling stable real-time pose tracking during subsequent data collection.

\textbf{Feasibility validation for robot execution.}
During data collection, the recorded gripper poses are simultaneously mapped to target robot poses and checked online for executability. Geometric reachability alone does not guarantee stable execution. Compared with offline inverse-kinematics checks, online validation during acquisition better captures execution-level constraints beyond static reachability, including inverse-solution failure, soft-limit violation, overspeed motion, and runtime communication anomalies. Unlike post-hoc replay screening, this allows infeasible motions to be identified during acquisition and improves collection efficiency.

\textbf{AR-based recovery mode.} 
The pipeline also supports recovery-oriented data collection through human intervention during policy execution. In the recovery mode, we use Pico 4 Ultra to provide real-time tactile feedback via augmented reality (AR), as shown in~\cref{fig:datacollect}. Our tAmeR system streams tactile videos together with wrist-view RGB fisheye videos to the headset, allowing the operator to observe rich contact cues and an egocentric, unobstructed view during teleoperation. tAmeR supports multiple tactile sensors, can be deployed across different robot embodiments, and remains low-cost in practice (\$630 for Pico 4 Ultra).

\subsection{Pyramid-Structured Data Regime}
\label{sec:method-data}
      \begin{figure*}[!t]
      \centering
      \includegraphics[width=\linewidth]{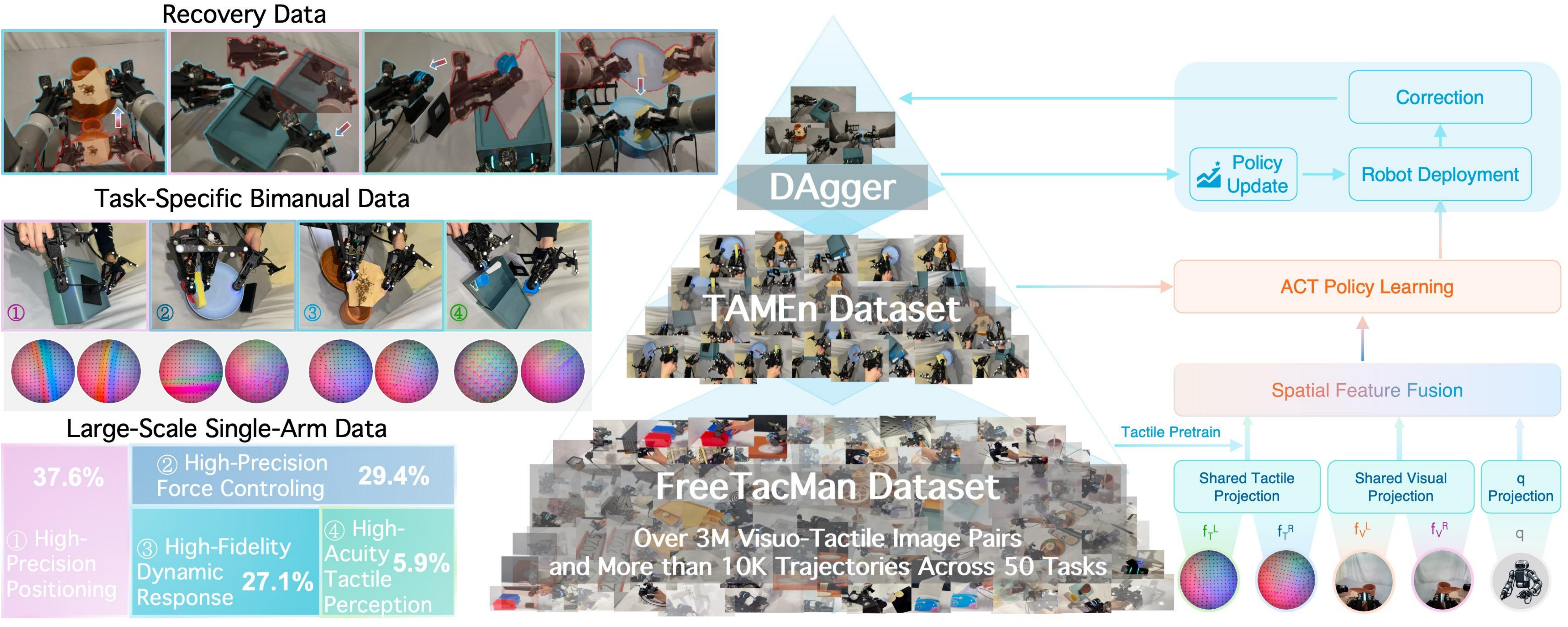}
      \caption{\textbf{Pyramid-structured visuo-tactile learning framework. }Large-scale single-arm visuo-tactile data provide broad priors for pretraining, task-specific bimanual data support coordination-aware fine-tuning, and recovery data further refine the policy around realistic failure states.
 }
      \label{fig:dataset}
   \end{figure*}
\methodname adopts a pyramid-structured data regime for staged training, as shown in~\cref{fig:dataset}. The base layer consists of large-scale single-arm visuo-tactile data, which are efficient to collect and provide broad priors over contact dynamics. In our implementation, we leverage FreeTacMan dataset~\cite{wu2025freetacman}, a large-scale multimodal dataset containing over 3000k paired visuo-tactile images with end-effector poses and 10k demonstration trajectories across 50 contact-rich manipulation tasks. This layer supports representation learning and policy initialization.
The middle layer consists of task-specific bimanual demonstrations, which adapt these tactile priors to coordinated manipulation. The top layer consists of recovery data collected from policy-induced failure cases, which further improve robustness under realistic failure modes.
This pyramid structure allows the policy to progress from broad visuo-tactile prior learning, to task-specific bimanual adaptation, and finally to recovery-oriented refinement.

\subsection{Closed-Loop Policy Learning}
\label{sec:method-policy}
Given the pyramid-structured data regime, we formulate downstream learning as a closed-loop visuo-tactile policy learning problem, as illustrated in~\cref{fig:dataset}. The policy takes two wrist-mounted fisheye RGB observations and two tactile observations as input, and predicts a 16-dimensional action comprising dual-arm joint commands and continuous gripper actions. Learning proceeds in three coupled stages that correspond to the three layers of the data pyramid. Large-scale single-arm visuo-tactile data are first used to pretrain tactile representations that capture broad contact dynamics. Task-specific bimanual demonstrations then provide supervision for learning coordinated contact-rich behaviors. Finally, recovery data collected around policy-induced failure states are incorporated through a DAgger-style update loop, allowing the policy to gradually adapt to the state distribution induced by its own execution. In this way, the pyramid-structured dataset is not only an organizational abstraction over data sources, but also a staged training regime that supports initialization, task adaptation, and iterative policy improvement.

To initialize the tactile branch, we pretrain tactile representations on the large-scale single-arm dataset using a contrastive objective. For each tactile embedding $\mathbf{t}_i$, we define a positive visual set $\mathcal{P}_i=\{\mathbf{v}_i,\mathbf{v}_{i+1}\}$ consisting of the aligned visual embedding at the current timestep and an additional temporal positive from the next timestep. The loss is written as:
\begin{equation}
\small
\begin{aligned}
\mathcal{L}_{\mathrm{con}}
=
-\frac{1}{B}\sum_{i=1}^{B}
\log
\frac{
\sum_{\mathbf{v}\in\mathcal{P}_i}
\exp(\mathbf{v}^{\top}\mathbf{t}_i/\tau)
}{
\sum_{\mathbf{v}\in\mathcal{P}_i}
\exp(\mathbf{v}^{\top}\mathbf{t}_i/\tau)
+
\sum_{\mathbf{v}\in\mathcal{N}_i}
\exp(\mathbf{v}^{\top}\mathbf{t}_i/\tau)
}
\end{aligned}
\end{equation}
where $B$ is the batch size, $\tau$ is the temperature parameter, and $\mathcal{N}_i$ denotes the set of negatives.

The downstream ACT policy is trained on task-specific bimanual demonstrations with a supervised action loss:
\begin{equation}
\mathcal{L}_{\mathrm{act}}
=
\sum_{i=1}^{T}
\left\|
\hat{\mathbf{a}}_i-\mathbf{a}_i
\right\|_1,
\end{equation}
where $\hat{\mathbf{a}}_i\in\mathbb{R}^{16}$ and $\mathbf{a}_i\in\mathbb{R}^{16}$ denote the predicted and demonstrated 16-dimensional actions at timestep $i$. After initial training, the policy is deployed on the robot, and human corrections are collected when execution enters failure-prone states. These corrected trajectories are added to the recovery set and incorporated into subsequent policy updates, allowing the policy to better match the state distribution induced by its own execution.

\section{Experiments}
We design experiments to address three key questions: 

\textbf{Q1}. Can the proposed data acquisition system improve the efficiency and quality of bimanual visuo-tactile data collection?

\textbf{Q2}. How much does tactile sensing improve bimanual manipulation performance?

\textbf{Q3}. To what extent do pretraining and recovery data improve robustness and generalization?

\subsection{Experimental Setup}
\label{sec:experiment-task}
\begin{figure}[!t]
      \centering
      \includegraphics[width=\linewidth]{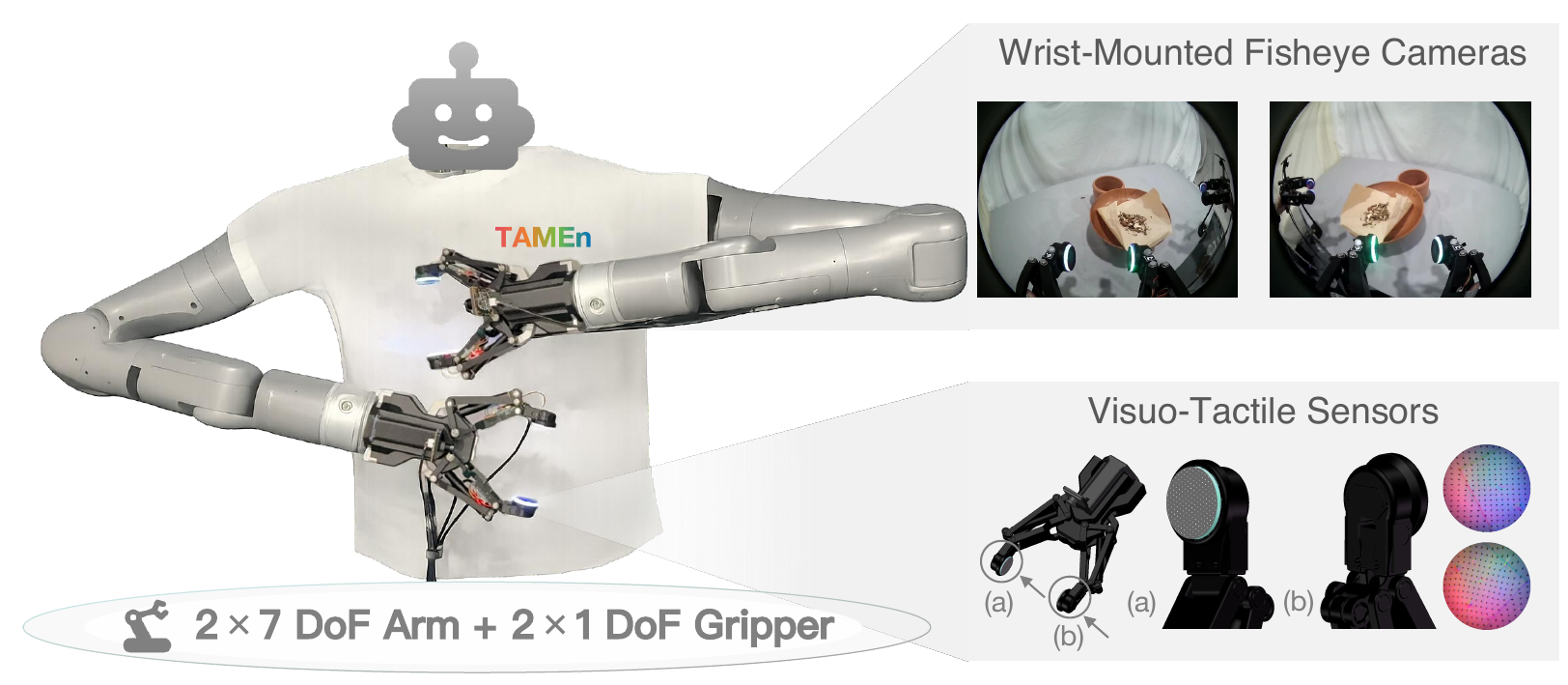}
      \caption{\textbf{Robot setup. }A dual-arm platform equipped with wrist-mounted cameras and fingertip visuo-tactile sensors.
 }
      \label{fig:setup}
   \end{figure}

To validate the real-world applicability of our approach, we deploy our policy
on a dual-arm JAKA K1 platform, as shown in~\cref{fig:setup}. The robotic system features a continuous 16-DoF action space,
driven by two 7-DoF arms and two DH grippers. It is equipped with two wrist-mounted fisheye cameras for visual observations and four fingertip visuo-tactile sensors for contact-rich perception. 
   
\begin{figure*}[t]
      \centering
      \includegraphics[width=\linewidth]{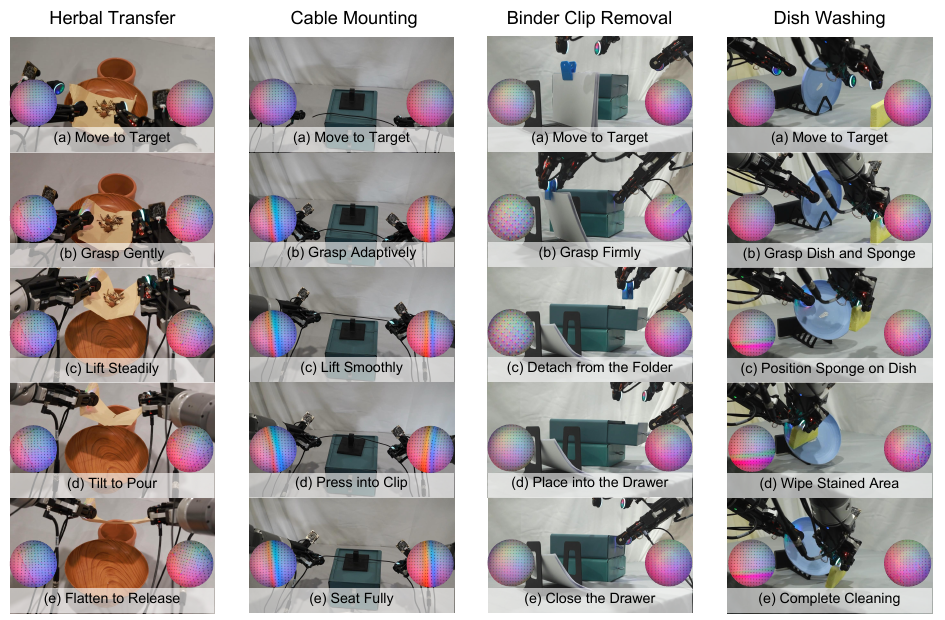}
      \caption{\textbf{Trajectory visualization.} We test \methodname on a variety of contact-rich tasks. 
      }
      \label{fig:dualarm_trajectory_visualization}
   \end{figure*}

To address these research questions, we conduct a comprehensive experimental suite covering deformable object manipulation (herbal transfer), high-precision alignment (cable mounting), multi-stage bimanual coordination (binder clip removal), and sustained contact control (dish washing), as illustrated in~\cref{fig:dualarm_trajectory_visualization}. 
1) \textbf{Herbal transfer}. The robot cooperatively manipulates a flexible sheet to lift the herbs and pour them into a target container. Successful execution requires stable bimanual coordination, careful handling of the deformable support, and precise control of tilting and release.
2) \textbf{Cable mounting}. The robot lifts a flexible cable, aligning it with a target clip, and pressing it into place. Successful execution requires adaptive grasping to prevent failure caused by cable motion during gripper closure, and contact-aware insertion to confirm successful seating when visual cues are unreliable.
3) \textbf{Binder clip removal}. The left arm grasps a binder clip attached to a folder, detaches it, and moves it toward the drawer, while the right arm opens the drawer and closes it after the clip is placed inside. Successful execution requires firm yet controlled interaction with the spring-loaded clip, as tactile feedback helps stabilize the grasp and judge detachment, followed by coordinated sequential manipulation for drawer opening, placement, and closing.
4) \textbf{Dish washing}. The robot grasps a dish and a sponge, positions the sponge onto the stained area, and wipes the surface until the stain is removed. Successful execution requires coordinated dual-object manipulation, stable contact between the sponge and the dish surface, and controlled wiping under sustained contact.

\subsection{User Study on Data Collection System}

\textbf{Tracking robustness (Q1).}
Reliable motion capture remains challenging in contact-rich manipulation due to occlusion, environmental interference, and sensitivity to operator experience, often resulting in unstable marker observations~\cite{qian2025openmocaprethinkingopticalmotion}. Object-based tracking can improve robustness under such practical conditions by incorporating structural constraints beyond individual marker detections. To assess this, we compare marker-only tracking and object-based tracking on the dish washing task across 10 users, evenly divided between experienced and novice participants. Each user performs 10 trials per setting, and success is defined as maintaining correct marker identities throughout the sequence.
Table~\ref{tab:tracking_robustness_userstudy} shows that marker-only tracking is substantially more sensitive to capture disturbances than object-based tracking. For novice users, failures are often associated with dropped markers, noisy observations, and unstable identity assignment. Even for experienced users, marker-only tracking can still fail under brief occlusions. In contrast, object-based tracking remains stable across users and capture conditions.
\begin{table}[t]
  \centering
  \caption{
    \textbf{Tracking robustness on the dish washing task.} The table reports tracking success rates (\%).
  }
  \label{tab:tracking_robustness_userstudy}
  \setlength{\tabcolsep}{4pt}  
  \scalebox{1}{
  \begin{tabular}{lccc}
    \toprule
    \textbf{Method} 
      & \makecell{Novice Users} 
      & \makecell{Experienced Users} 
      & \cellcolor[gray]{0.9}\textbf{Avg.} \\
    \midrule
    \makecell{Marker-only Tracking} 
      & 32  & 78 & \cellcolor[gray]{0.9}55  \\
    \makecell{Object-Based Tracking (Ours)} 
      & 100 & 100 & \cellcolor[gray]{0.9}100  \\
    \bottomrule
  \end{tabular}
  }
\end{table}

\begin{figure}[t]
      \centering
      \includegraphics[width=\linewidth]{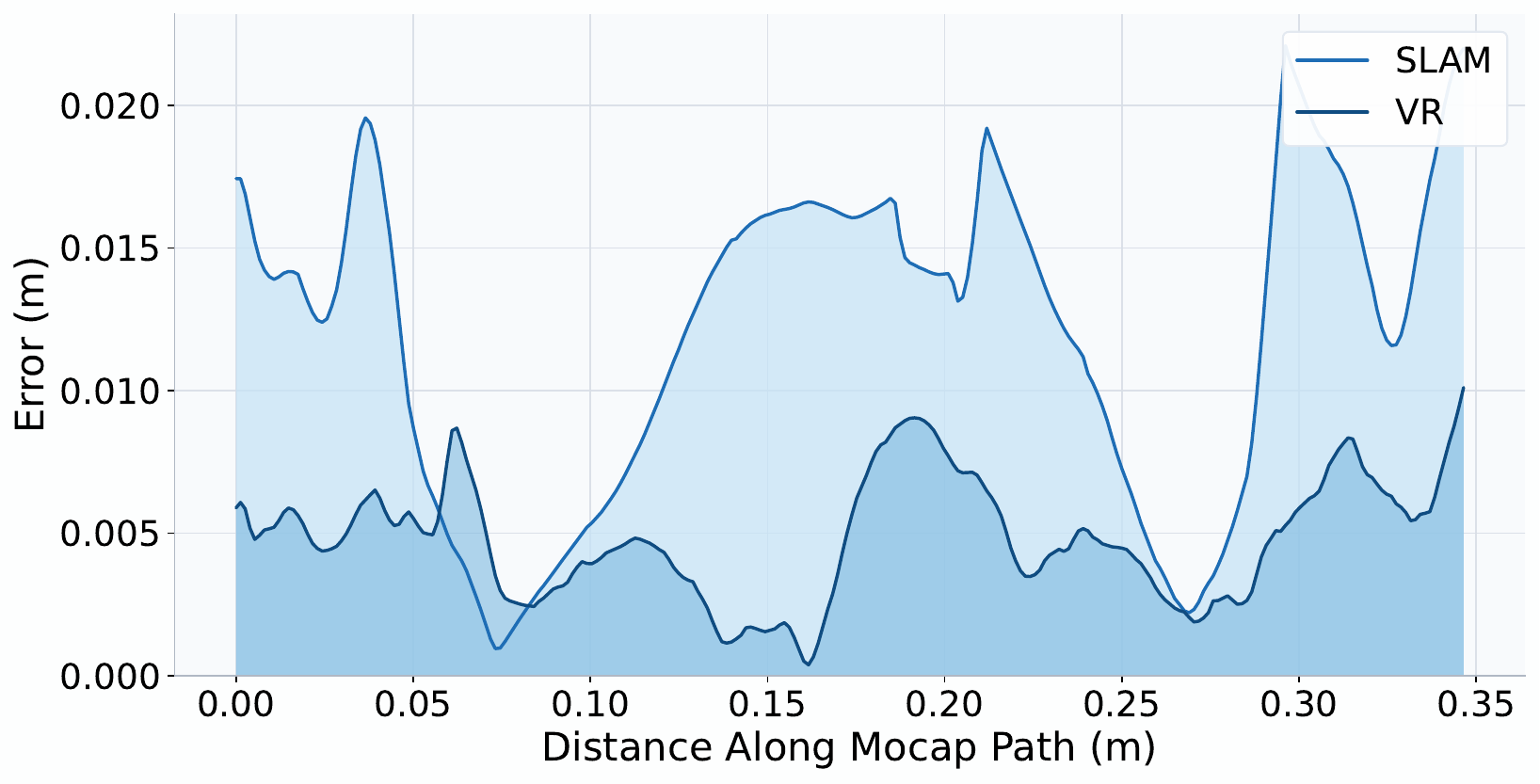}
      \caption{\textbf{Tracking accuracy.} Trajectory errors of VR-based tracking and GoPro-based SLAM tracking relative to the motion-capture reference. 
      }
      \label{fig:tracking_error}
   \end{figure}

\textbf{Tracking accuracy (Q1).}
Accurate and portable pose tracking is critical for practical visuo-tactile data collection, yet existing solutions involve trade-offs between precision, robustness, and deployment flexibility. As shown in~\cref{fig:tracking_error}, we further compare the trajectories obtained from high-precision motion capture, VR-based tracking, and GoPro-based tracking. Motion capture provides a highly reliable reference with sub-millimeter precision, but its reliance on external base stations limits portability. SLAM-based tracking~\cite{chi2024universal,yin2026rapidreconfigurableadaptiveplatform} offers a portable alternative, yet its error increases in low-feature environments (\textit{e.g.}, drawer opening). VR-based tracking mitigates these limitations by providing a portable solution with improved stability, while maintaining trajectory errors within 1\,cm.

\textbf{Data validity (Q1).}
Human demonstrations are not always directly executable on the robot, as motions that appear natural during collection may still violate robot-level execution constraints.
To evaluate the effect of the proposed online validation mechanism, we compare two data collection settings, one without feasibility screening and the other with online validation.
The study is conducted on the herbal transfer and cable mounting tasks with 10 users of varying experience levels. For each setting and each task, 10 trajectories are collected and evaluated by whether they can be successfully replayed on the robot.
As shown in~\cref{tab:replayability_userstudy}, online validation during collection significantly improves data validity. Without feasibility screening, the replay success rates are only 39\% for herbal transfer and 12\% for cable mounting. For herbal transfer, rapid shaking during pouring can lead to unstable robot motions. For cable mounting, lifting the cable too high may exceed the robot workspace. By filtering out such infeasible motions during acquisition, online validation leads to 100\% replay success on both tasks.
\begin{table}[t]
  \centering
  \caption{
    \textbf{Data validity across collection settings.} The table reports replay success rates (\%).
  }
  \label{tab:replayability_userstudy}
  \setlength{\tabcolsep}{4pt}  
  \scalebox{1}{
  \begin{tabular}{lccc}
    \toprule
    \textbf{Method} 
      & \makecell{Herbal\\Transfer} 
      & \makecell{Cable\\Mounting} 
      & \cellcolor[gray]{0.9}\textbf{Avg.} \\
    \midrule
    \makecell{No Feasibility Screening} 
      & 39 & 12 & \cellcolor[gray]{0.9}26 \\
    \makecell{Online Validation (Ours)} 
      & 100 & 100 & \cellcolor[gray]{0.9}100 \\
    \bottomrule
  \end{tabular}
  }
\end{table}

\subsection{Validation on Downstream Policy Learning}
We evaluate how different layers of the pyramid-structured data regime contribute to downstream imitation learning. Each task in~\cref{fig:dualarm_trajectory_visualization} is trained and evaluated over 20 trials.

\begin{itemize} [leftmargin=*,itemsep=0pt,topsep=0pt]
    \item \textbf{ACT}~\cite{zhao2023act} \textbf{(Vision-Only)}: The original ACT policy uses only RGB observations from the wrist-mounted fisheye camera.
    \item \textbf{Ours (+ Tactile w/o Pretrain)}: A visuo-tactile extension of ACT that takes both visual and tactile observations as input, where the two modalities are encoded by identical backbone architectures trained from scratch.
    \item \textbf{Ours (+ Pretrain)}: We further initialize the tactile encoder using the proposed multi-positive contrastive pretraining objective with both primary and secondary positives. For this large-scale single-arm data layer, we leverage FreeTacMan~\cite{wu2025freetacman}, a sub-millimeter-precision visuo-tactile manipulation dataset containing over 3M visuo-tactile image pairs and more than 10K trajectories across 50 tasks.
    \item \textbf{Ours (+ Pretrain + DAgger)}: It further augments the pretrained visuo-tactile policy with recovery data collected through human correction, and applies DAgger-style policy updates to improve robustness under policy-induced failures.
\end{itemize}

\textbf{Tactile input provides direct contact cues that are often unavailable from vision alone (Q2).}
As shown in~\cref{tab:tac}, incorporating tactile sensing improves the average success rate from 34\% to 55\%. For example, in cable mounting, tactile feedback helps detect whether the deformable cable slips out of the gripper during closure and whether it has been successfully seated in the clip despite the low visual contrast between them. Similarly, in dish washing, tactile feedback helps determine whether the sponge has made and maintained stable contact with the dish surface throughout wiping.

\textbf{Tactile pretraining on the large-scale dataset provides further gains beyond tactile input alone (Q3).} 
 As shown in~\cref{tab:tac}, tactile pretraining further improves the average success rate from 55\% to 65\%. 
 Since all evaluated tasks involve multi-stage manipulation, successful execution requires the policy to remain stable across sequential subgoals and adapt to changing contact conditions.
 The improvement is particularly evident in herbal transfer and binder clip removal. In herbal transfer, the robot must maintain stable support of a deformable object while coordinating lifting, tilting, and release. This is especially important during the final stage of pouring, where the remaining material must be released by flattening the paper without tearing it.
In binder clip removal, tactile feedback helps the robot maintain a stable grasp while adapting to the changing resistance of the spring-loaded clip during removal. These results suggest that tactile pretraining improves the policy’s ability to handle stage-dependent tactile variations and maintain robust behavior throughout multi-stage manipulation.
\begin{table}[t]
  \centering
  \caption{
    \textbf{Policy success rates (\%) across tasks.} Tactile input and pretraining significantly boost imitation learning.
  }
  \label{tab:tac}
  \setlength{\tabcolsep}{3.5pt}  
  \scalebox{1}{
  \begin{tabular}{lccccc}
    \toprule
    \textbf{Method} 
      & \makecell{Herbal\\Transfer} & \makecell{Cable\\Mounting} & \makecell{Binder Clip\\Removal} & \makecell{Dish\\Washing} & \cellcolor[gray]{0.9}\textbf{Avg.} \\
    \midrule
    \makecell{ACT~\cite{zhao2023act}\\(Vision-Only)}      & 40 & 10 & 50 & 35 & \cellcolor[gray]{0.9}34 \\
    \makecell{Ours (+ Tactile \\w/o Pretrain)}          & 65 & 30 & 65 & 60 & \cellcolor[gray]{0.9}55 \\
    \makecell{Ours\\(+ Pretrain)}       & \textbf{75} & \textbf{40} & \textbf{80} & \textbf{65} & \cellcolor[gray]{0.9}\textbf{65} \\
    \bottomrule
  \end{tabular}
  }
\end{table}

\textbf{Recovery data further improve the policy by enabling it to recover from failure-prone states during execution (Q3).}
As presented in Table \cref{tab:dagger}, adding only 10\% online recovery data increases the average success rate from 65\% to 75\%. These online corrections directly target realistic failure scenarios. In cable mounting, they include grasp readjustment during cable pickup, sustained downward pulling during insertion, and position correction after an initially misaligned press. In binder clip removal, the grasping position can be adjusted when the initial approach is misaligned. In dish washing, the wiping motion can be retried when the first attempt fails to make effective contact. Such targeted interventions are highly data efficient. In contrast, adding 50\% additional nominal demonstrations yields only 70\% success, indicating that simply scaling up normal data is less effective than focused online recovery. Furthermore, offline recovery data collected by directly imitating near-failure states using the handheld gripper achieve only 56\% success. These offline demonstrations lack real execution context and fall out of distribution, thus offering less corrective value than online recovery data.

\begin{table}[t]
  \centering
  \caption{
    \textbf{Policy success rates (\%) across tasks.} Effect of recovery-based data aggregation on imitation learning.
  }
  \label{tab:dagger}
  \setlength{\tabcolsep}{3.5pt}  
  \scalebox{1}{
  \small
  \setlength{\tabcolsep}{2pt}
  \begin{tabular}{lccccc}
    \toprule
    \textbf{Method} 
       & \makecell{Herbal\\Trans.} & \makecell{Cable\\Mount.} & \makecell{Binder Clip\\Rem.} & \makecell{Dish\\Wash.} & \cellcolor[gray]{0.9}\textbf{Avg.} \\
    \midrule
    \makecell{Ours\\(+ Pretrain)}      & 75  & 40 & 80 & 65 & \cellcolor[gray]{0.9}65 \\
    \makecell{Ours\\(+ Pretrain \\+ 50\% Nominal)}      & 80  & 50 & 80 & 70 & \cellcolor[gray]{0.9}70 \\
    \makecell{Ours\\(+ Pretrain \\+ 10\% Offline Recovery)}      & 50  & 40 & 70 & 65 & \cellcolor[gray]{0.9}56 \\
    \makecell{Ours\\(+ Pretrain \\+ 10\% Online Recovery)}        & \textbf{80} & \textbf{50} & \textbf{90} & \textbf{80} & \cellcolor[gray]{0.9}\textbf{75} \\
    \bottomrule
  \end{tabular}
  }
\end{table}

\textbf{\methodname generalizes to unseen objects (Q3).}
   \begin{figure*}[!t]
      \centering
      \includegraphics[width=\linewidth]{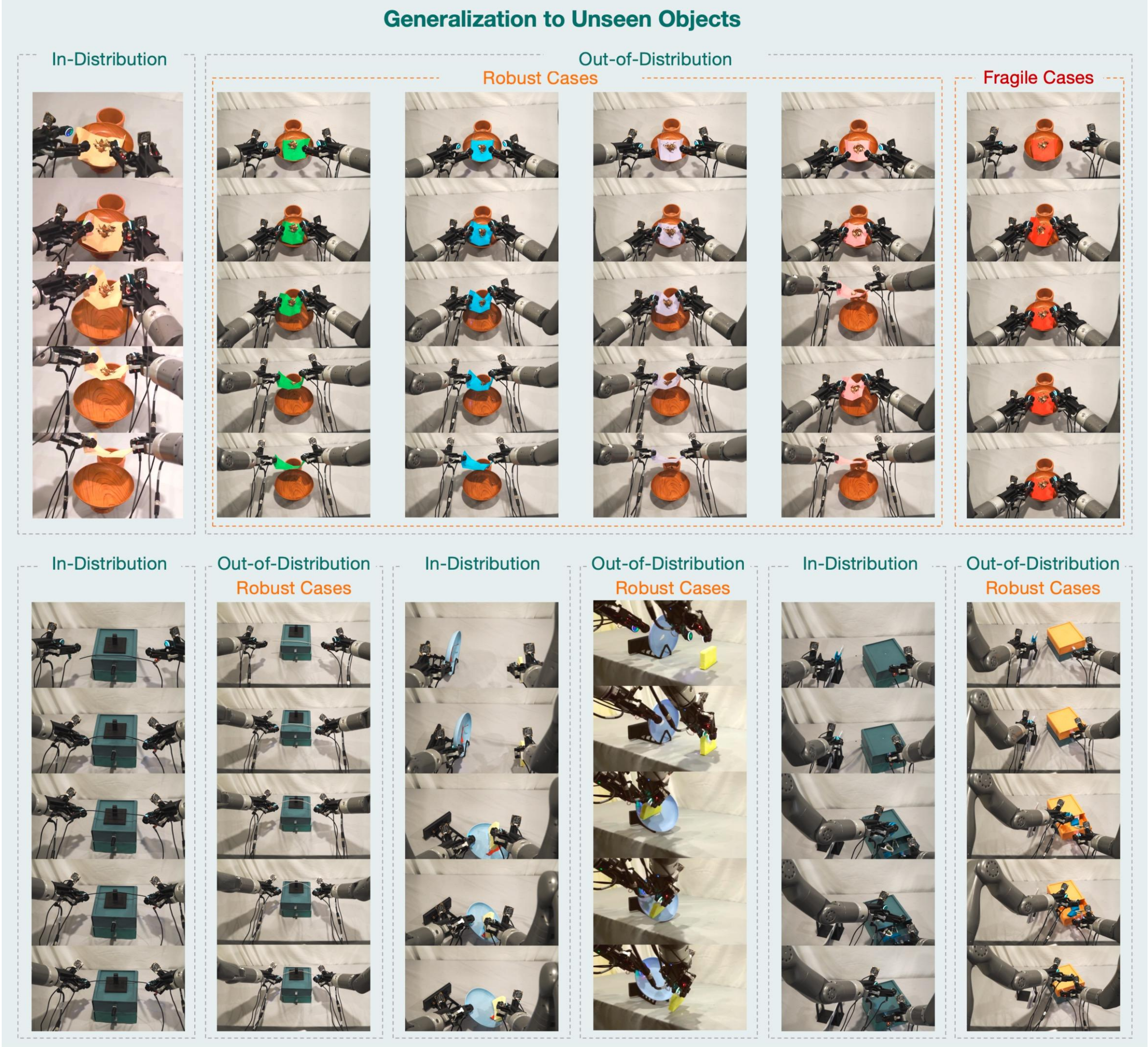}
      \caption{\textbf{Generalization.}~\methodname generalizes to unseen objects with representative robust and fragile cases shown.
 }
      \label{fig:generalization}
   \end{figure*}
As shown in~\cref{fig:generalization}, we evaluate unseen-object generalization on all tasks by altering the appearance of the evaluated objects. Specifically, herbal transfer uses five papers with unseen colors, binder clip removal changes the drawer color from blue to yellow, and cable mounting replaces the black cable with white cable. For dish washing task, to avoid material waste, data collection is performed using strip-based props with different colors, whereas evaluation in~\cref{tab:tac} and~\cref{tab:dagger} is conducted using white jam, as shown in~\cref{fig:dishwashingtask}. These changes preserve the underlying task structure while altering visual appearance, thereby testing whether the policy can rely on visuo-tactile cues rather than overfitting to the original object textures and colors. 
\Cref{tab:generalization} shows that \methodname retains clear advantages on unseen objects. The vision-only policy fails almost completely in herbal transfer and cable mounting, whereas \methodname still reaches 60\% and 30\% success, with only small drops in key contact-rich stages. A similar trend is observed in binder clip removal, where whole-task success improves from 30\% to 60\%. These results suggest that tactile pretraining and recovery data improve policy transfer across object variations.

   \begin{figure*}[!t]
      \centering
      \includegraphics[width=\linewidth]{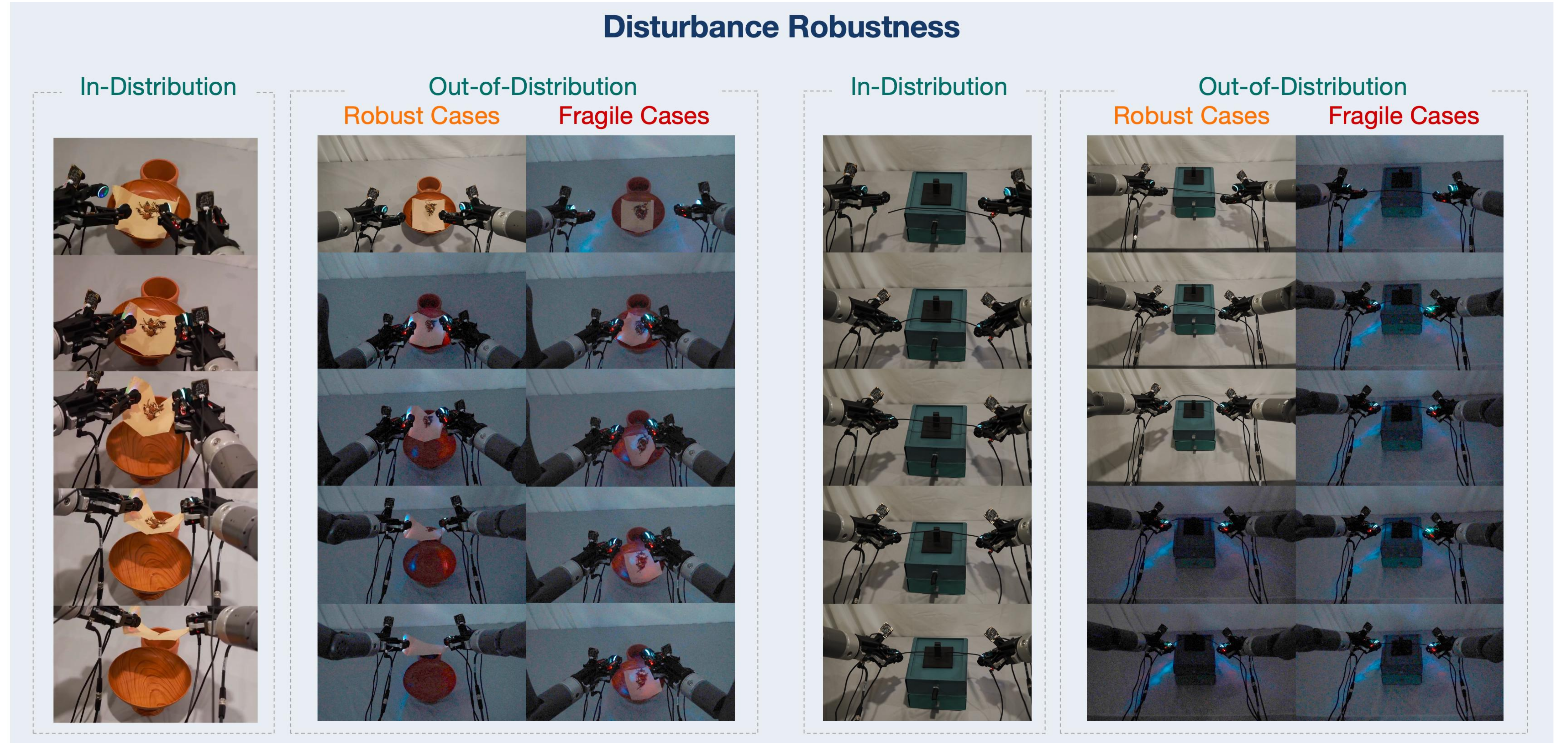}
\caption{\textbf{Robustness.}~Under visual disturbances, \methodname exhibits improved robustness during contact-rich execution, with representative robust and fragile cases shown.
}
      \label{fig:robustness}
   \end{figure*}
\textbf{\methodname remains robust under external disturbances (Q3).}
We further test robustness to lighting disturbances in herbal transfer and cable mounting by switching the illumination during execution, as shown in~\cref{fig:robustness}. We focus on these two tasks because their critical stages, namely pouring and insertion, are highly contact-rich and therefore provide a clear test of whether tactile sensing can compensate when visual observations become unreliable.
We consider two disturbance settings. In the full disturbance setting, illumination changes throughout the whole episode. In the post-grasp disturbance setting, the perturbation is introduced only after the object has already been grasped. This design separates failures caused by visual degradation during object acquisition from those during subsequent contact-rich interaction.
  As shown in~\cref{tab:robustness}, the strongest gains appear in the post-grasp setting, suggesting that tactile pretraining and recovery data are particularly helpful once contact-rich execution begins. 
In herbal transfer, both methods fail under full disturbance, indicating that object acquisition remains challenging when severe visual degradation is introduced during the approach stage.
Under post-grasp disturbance, however,~\methodname consistently succeeds in the subsequent pouring stage. In cable mounting, although cable localization still depends partly on vision,~\methodname nevertheless achieves 30\% whole-task success under full disturbance and 40\% under post-grasp disturbance, while the vision-only policy fails completely in both settings.
Overall, these results show that tactile pretraining and recovery data substantially improve robustness when visual perception is degraded, especially during contact-rich execution.

\begin{table}[t]
\caption{\textbf{Generalization to unseen objects.}
Visuo-tactile learning with tactile pretraining and DAgger significantly improves performance on unseen objects.}
\label{tab:generalization}
\centering
\small 
\begin{tabular}{llc>{\columncolor{gray!15}}c}  
\toprule
\textbf{Task} & \textbf{Phase} & \textbf{\makecell{Ours \\(Vision-Only)}} & \textbf{\makecell{Ours (+ Pretrain\\+ DAgger)}} \\
\midrule  
\multirow{3}{*}{\makecell[l]{Herbal\\Transfer}} 
 & Grasp       & 0\% & \textbf{70\%} \\  
 & Pour        & 0\% & \textbf{60\%} \\
 & Whole Task  & 0\% & \textbf{60\%} \\
\midrule
\multirow{3}{*}{\makecell[l]{Cable\\Mounting}} 
 & Pick       & 0\% & \textbf{40\%} \\
 & Seat Fully  & 0\% & \textbf{30\%} \\
 & Whole Task  & 0\% & \textbf{30\%} \\
\midrule
\multirow{3}{*}{\makecell[l]{Binder Clip\\Removal}} 
 & Detach       & 40\% & \textbf{70\%} \\
 & Open Drawer  & 30\% & \textbf{60\%} \\
 & Whole Task  & 30\% & \textbf{60\%} \\
\bottomrule
\end{tabular}
\end{table}

\begin{table}[t]
\caption{\textbf{Robustness in disturbed conditions.}
Tactile pretrain and DAgger improve robustness in contact-rich stages. Dist. denotes disturbance.}
\label{tab:robustness}
\centering
\small 
\begin{tabular}{llcccc}
\toprule
\textbf{Task} & \textbf{Phase} & \multicolumn{2}{c}{\textbf{\makecell{Ours\\(Vision-Only)}}} & \multicolumn{2}{c}{\textbf{\makecell{Ours(+ Pretrain\\+ DAgger)}}} \\
\cmidrule(lr){3-4} \cmidrule(lr){5-6}
 &  & \makecell{Full\\Dist.} & \makecell{Post-Grasp\\Dist.} & \makecell{Full\\Dist.} & \cellcolor[gray]{0.9}\makecell{Post-Grasp\\Dist.} \\
\midrule
\multirow{3}{*}{\makecell[l]{Herbal\\Transfer}} 
 & Grasp       & 0\% & \textbf{20\%} & 0\% & \cellcolor[gray]{0.9}\textbf{70\%} \\
 & Pour        & 0\% & \textbf{10\%} & 0\% & \cellcolor[gray]{0.9}\textbf{70\%} \\
 &Whole Task   & 0\% & \textbf{10\%} & 0\% & \cellcolor[gray]{0.9}\textbf{70\%} \\
\midrule
\multirow{3}{*}{\makecell[l]{Cable\\Mounting}} 
 & Pick        & 0\% & \textbf{0\%} & 60\% & \cellcolor[gray]{0.9}\textbf{90\%} \\
 & Seat Fully  & 0\% & \textbf{0\%} & 30\% & \cellcolor[gray]{0.9}\textbf{40\%} \\
 & Whole Task  & 0\% & \textbf{0\%} & 30\% & \cellcolor[gray]{0.9}\textbf{40\%} \\
\bottomrule
\end{tabular}
\end{table}

\section{Conclusion}
We present \methodname, a tactile-aware manipulation engine for closed-loop data collection in bimanual contact-rich tasks. The system integrates an adaptive dual-mode visuo-tactile acquisition pipeline that supports both high-precision motion capture and portable VR-based tracking. A feasibility-aware validation mechanism, combined with a pyramid-structured data regime, ensures that collected demonstrations are reliably replayable on robots and organizes heterogeneous data for staged learning. Furthermore, \methodname establishes a closed-loop data flywheel through AR-based teleoperation with tactile feedback (tAmeR), enabling recovery-oriented data collection and continuous policy refinement under realistic failure states. Experimental results show that our framework significantly improves replayability and policy learning, highlighting the value of visuo-tactile input, large-scale pretraining, and recovery-based refinement for robust contact-rich manipulation.

\textbf{Limitation and future work.} 
Despite the encouraging results, several extensions remain to be explored. While our current system has been validated on visuo-tactile grippers, an important next step is to extend the framework to dexterous hands, enabling finer-grained manipulation in more complex scenarios. In addition, while the proposed framework already accommodates multiple tactile sensors, a promising direction is to evaluate cross-sensor generalization beyond hardware compatibility. Specifically, it is important to examine whether data collected using one sensor can support learning on another, and whether policies trained with one sensing modality can transfer across sensors with only minor adaptation.

\section*{Acknowledgments}
We gratefully acknowledge Qianyu Guo, Checheng Yu, Chonghao Sima, Jingmin Zhang, and Chenyu Lin for their valuable insights and constructive discussions. We also extend our sincere gratitude to JAKA for their generous hardware and technical support.

{
\small
\bibliographystyle{IEEEtran}
\bibliography{bibliography_short, refs}  %
}

\clearpage
\appendices
\appendix[]

\subsection{Hardware Design}
\textcolor{gray}{\textit{| Supplement to \cref{sec:method-hardware} in the Main paper.}}

\begin{figure}[!b]
      \centering
      \includegraphics[width=\linewidth]{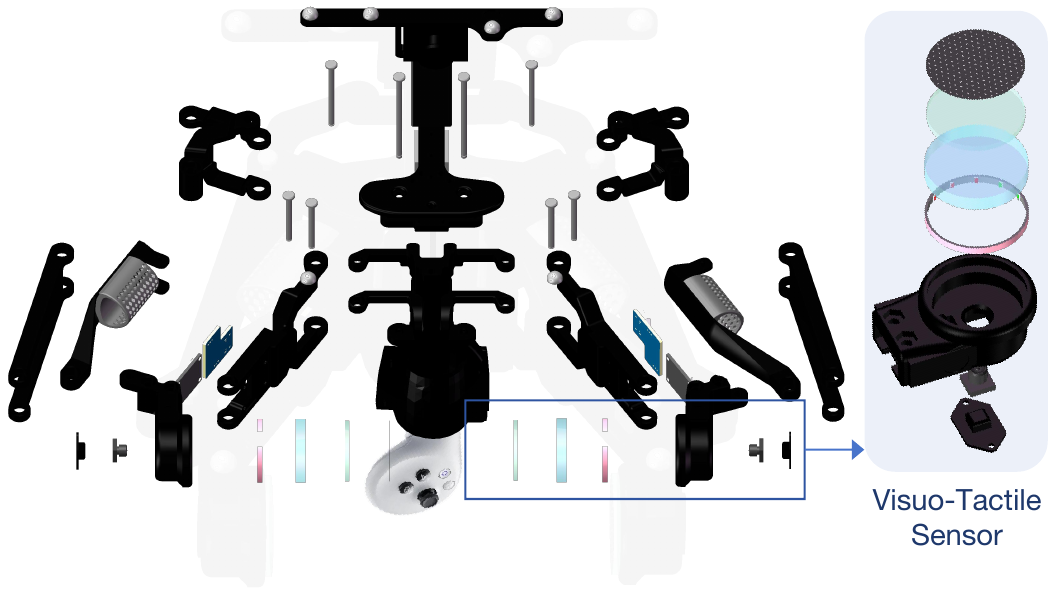}
      \caption{\textbf{Exploded view of~\methodname. }\textit{Left}: Overall interface structure. \textit{Right}: Exploded view of the visuo-tactile sensor.
 }
      \label{fig:explodedview}
   \end{figure}

\textbf{Manufacturing and assembly details.}
\Cref{fig:explodedview} shows the exploded view of \methodname. The structural components are fabricated using 3D printing, enabling rapid manufacture at low cost. The fingertip sleeves are fabricated via rigid–soft hybrid printing, improving comfort while preserving a compact form factor. Mechanical transmission components, such as bearings, screws, and nuts, are standard off-the-shelf parts, while the shafts are machined from 42CrMo steel for smooth and durable operation. The sensing module combines the camera, illumination system, elastomer, and supporting structures into an integrated assembly for tactile imaging. Users can flexibly equip the interface with motion-capture markers or a VR controller according to the operating mode.

\begin{figure}[!b]
      \centering
      \includegraphics[width=\linewidth]{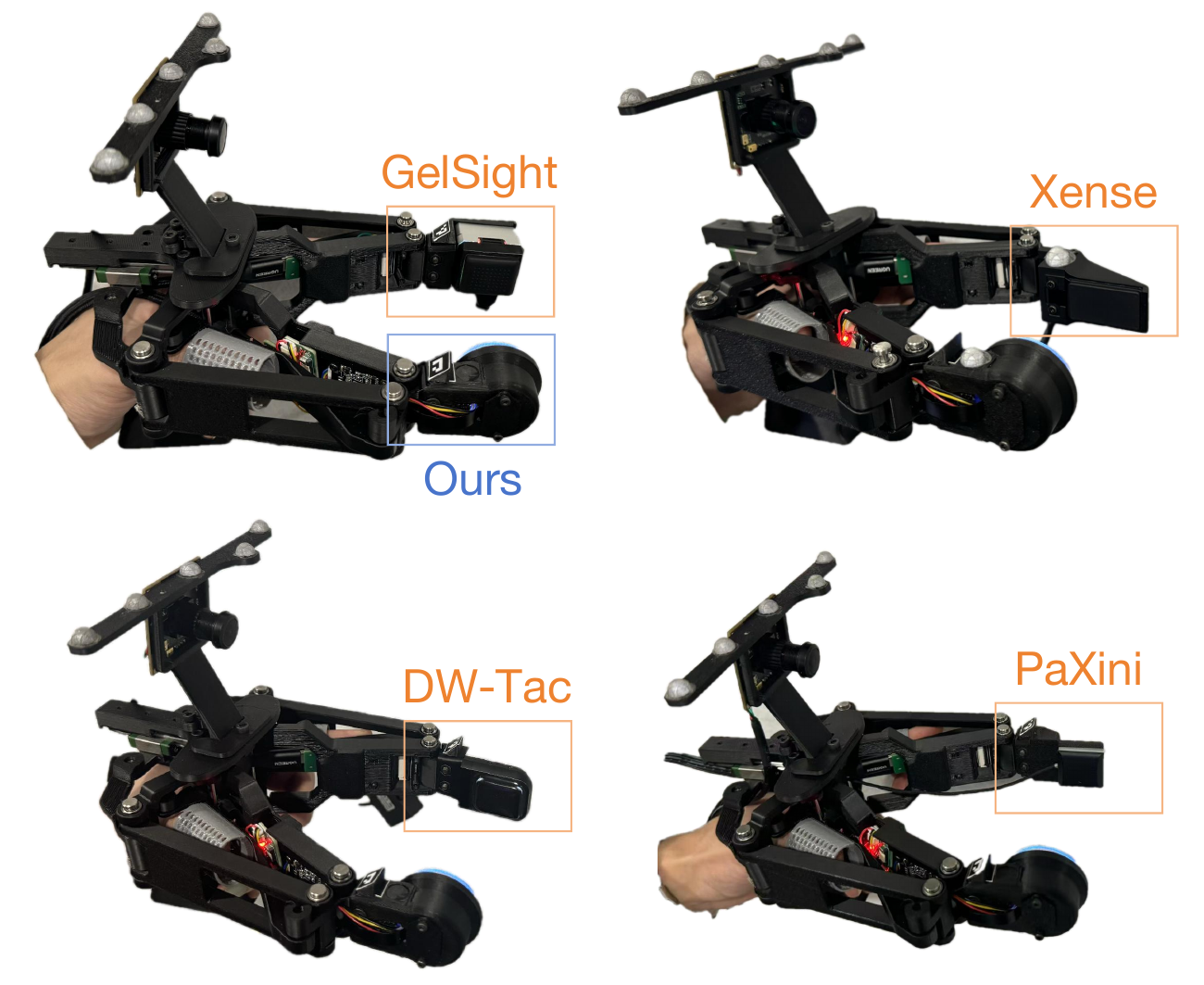}
      \caption{\textbf{Compatibility with multiple tactile sensors.}~\methodname supports seamless integration of different tactile sensors, including GelSight, Xense, DW-Tac, PaXini, and ours, demonstrating its adaptability across heterogeneous sensing modalities.
 }
      \label{fig:appendix_multisensor}
   \end{figure}

In addition to adapting to different gripper morphologies, the proposed handheld interface also supports multiple visuo-tactile sensors. As shown in~\cref{fig:appendix_multisensor}, GelSight, Xense, DW-Tac, PaXini, and our sensor can all be integrated into the same platform with only minor local modification. This compatibility is enabled by the shared mechanical backbone of the interface, which keeps the overall structure unchanged while allowing different fingertip sensing modules to be mounted in a modular manner. Such flexibility makes the platform easier to reproduce and extend, and also supports broader research on visuo-tactile data collection and downstream policy learning across different sensor choices. We will open-source the hardware models to facilitate reproduction and future development by the community.
   
\subsection{Data Collection}
\textcolor{gray}{\textit{| Supplement to \cref{sec:method-datacollect} in the Main paper.}}

\begin{figure}[!b]
      \centering
      \includegraphics[width=\linewidth]{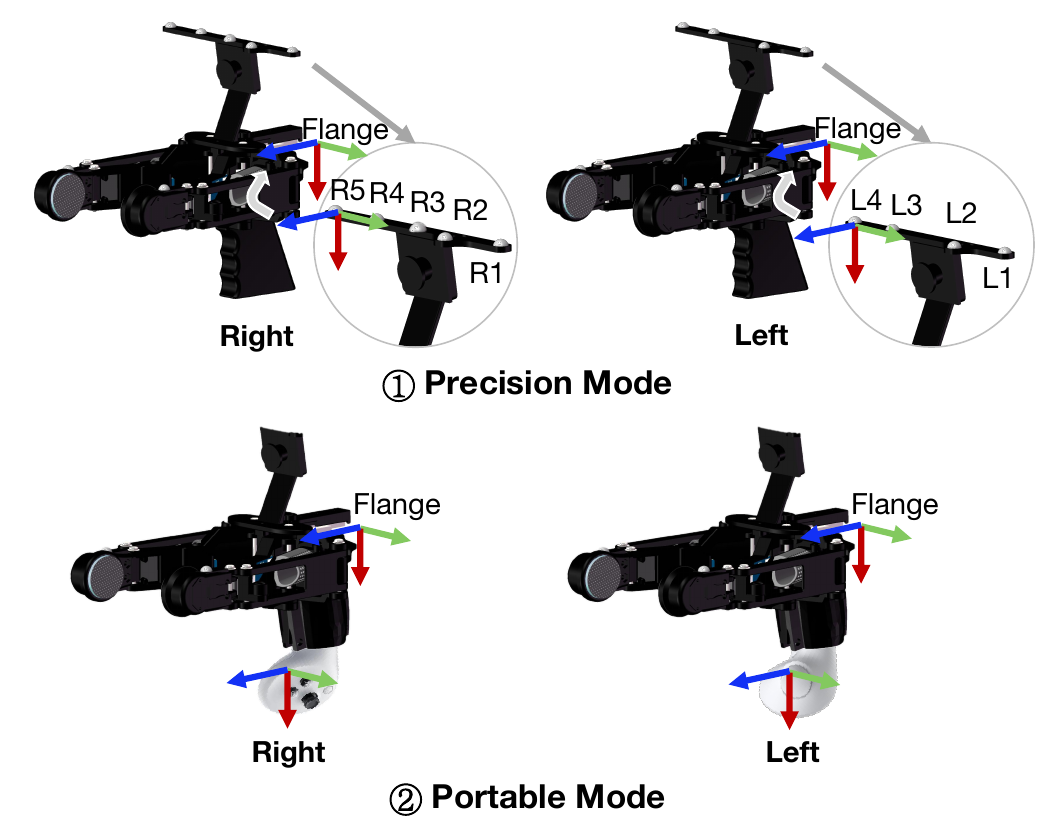}
      \caption{\textbf{Local frame construction and unified flange representation.} In the precision mode, a local frame is constructed from the marker configuration on each collector, while in the portable mode, the VR handle provides the tracked pose. Both are mapped to a shared flange-based reference frame for consistent pose representation.
 }
      \label{fig:appendix_coordinate}
   \end{figure}

\textbf{Human-to-robot data transfer}. To unify pose representations across the precision and portable setups, we define a shared end-effector reference frame on the flange of each handheld collector, as shown in~\cref{fig:appendix_coordinate}.
In the precision mode, the NOKOV motion capture system tracks a structured marker layout mounted on the collector. A local coordinate frame is then defined from the marker configuration. Specifically, the $y$-axis is determined by the two markers with the largest separation, namely $R_1$ and $R_5$ for the right collector, and $L_1$ and $L_4$ for the left collector. The $x$-axis is taken as the normal of the plane spanned by the markers, and the $z$-axis is obtained by enforcing a right-handed coordinate system orthogonal to the $x$- and $y$-axes.
In the portable mode, the VR system directly outputs the pose of the handle in its native tracking frame.
For both setups, the tracked pose is further mapped to the shared flange frame using a fixed geometric offset derived from the CAD model of the collector.

\textbf{Precise visuo-tactile data acquisition. }
In the precise configuration, we record synchronized visual, tactile, and motion data for high-fidelity demonstration collection. Visual observations are captured using a fisheye camera equipped with a $180^\circ$ field-of-view lens at 30 FPS and a resolution of $640 \times 480$. The visuo-tactile sensor integrates an RGB camera operating at the same frame rate and resolution, enabling temporally aligned multimodal observations. End-effector poses and gripper opening are tracked using the NOKOV motion capture system at 240~Hz, providing sub-millimeter precision for accurate trajectory recording. This configuration enables reliable trajectory capture for high-quality visuo-tactile data collection.

\begin{figure}[!t]
      \centering
      \includegraphics[width=\linewidth]{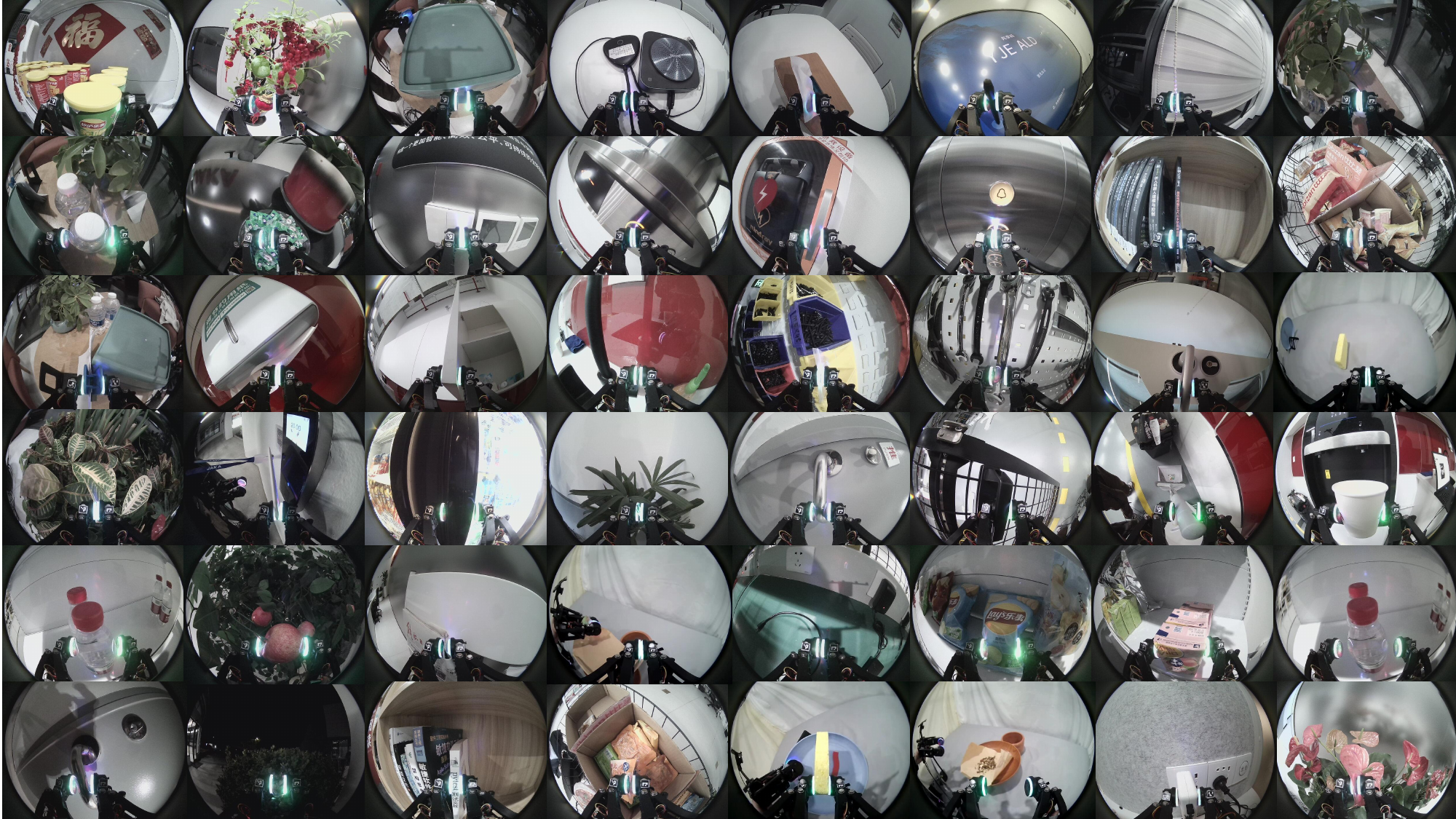}
      \caption{\textbf{In-the-wild visuo-tactile data collection. }The portable configuration of \methodname enables data acquisition across diverse real-world scenes.
 }
      \label{fig:collectinthewild}
   \end{figure}

\textbf{In-the-wild visuo-tactile data acquisition. }
In the portable configuration, we record synchronized visual, tactile, and motion data for collection in unstructured real-world environments, as shown in~\cref{fig:collectinthewild}. 
Visual and tactile observations are captured in the same way as in the precise configuration.
 End-effector poses are tracked using a portable VR system at 100~Hz. To improve tracking robustness, the VR controller is mounted with its sensing module facing the headset during operation. Gripper opening is tracked separately using ArUco markers attached to the gripper mechanism.

\begin{figure}[!t]
      \centering
      \includegraphics[width=\linewidth]{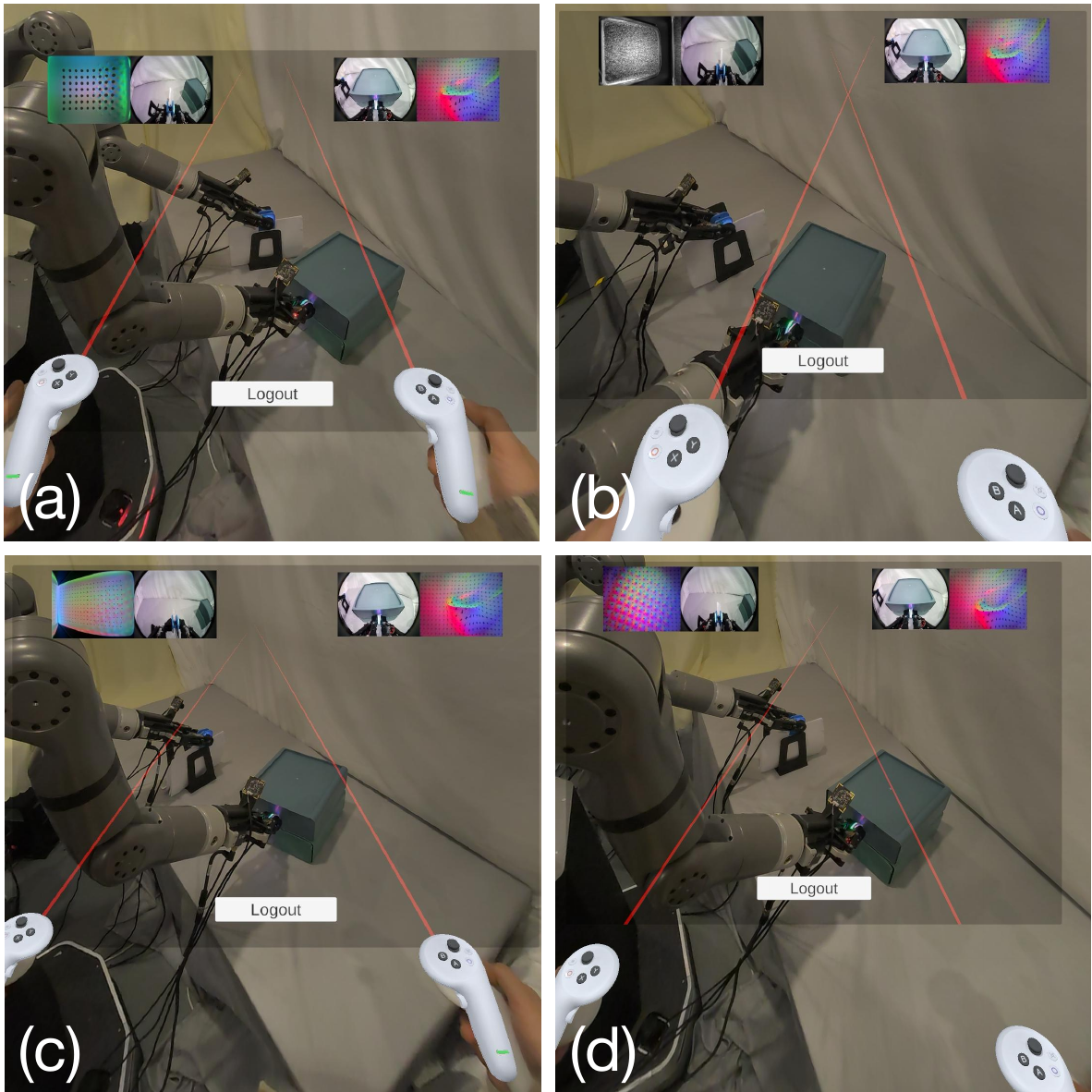}
      \caption{\textbf{Visualization of tAmeR. }Wrist-mounted RGB and tactile streams are visualized above the scene. The interface supports multiple visuo-tactile sensors, where (a) GelSight, (b) DW-Tac, (c) Xense, and (d) our sensor show the tactile stream from the left gripper.
 }
      \label{fig:4sensorsinvr}
   \end{figure}

\textbf{tAmeR for recovery-oriented data collection. }
We develop tAmeR, an AR-based application for immersive teleoperation.~\Cref{fig:4sensorsinvr} illustrates the visualization of tAmeR. It reconstructs the surrounding environment in mixed reality, allowing the operator to interact with the scene in an egocentric view. The interface streams wrist-mounted RGB observations and tactile images in real time. This design alleviates occlusion in conventional teleoperation and compensates for the lack of tactile feedback.

\textbf{Feasibility validation details.}
During data collection, the tracked human motion is mapped to robot targets and checked online for executability before being retained as valid demonstrations. The screening considers whether the mapped motion admits a valid inverse-kinematics solution, remains within the workspace and joint soft limits, and satisfies runtime motion constraints such as joint-speed and tcp-speed bounds. Motions that violate these conditions are identified during collection and trigger real-time feedback to the operator, allowing corrective adjustments.
In our implementation, the maximum joint velocity is set to $180^\circ/\mathrm{s}$, and the TCP velocity limit is set to $250,\mathrm{mm/s}$. The joint soft limits are defined as $[-360^\circ, 360^\circ]$ for J1, $[-105^\circ, 105^\circ]$ for J2, $[-360^\circ, 360^\circ]$ for J3, $[-145^\circ, 30^\circ]$ for J4, $[-360^\circ, 360^\circ]$ for J5, $[-105^\circ, 105^\circ]$ for J6, and $[-360^\circ, 360^\circ]$ for J7.

\subsection{Task and Evaluation}
\textcolor{gray}{\textit{| Supplement to \cref{sec:experiment-task} in the Main paper.}}

We evaluate \methodname on a dual-arm JAKA K1 platform equipped with two 7-DoF robot arms, DH grippers, wrist-mounted fisheye cameras, and fingertip visuo-tactile sensors. The policy outputs a continuous 16-DoF action, including dual-arm joint commands and gripper actions. All downstream policies are trained and evaluated on the four representative bimanual tasks shown in~\cref{fig:dualarm_trajectory_visualization}, covering deformable object handling, contact-aware insertion, sequential manipulation, and sustained contact control. Each task is evaluated over 20 real-robot trials.

\textbf{Herbal transfer.}
In this task, the robot uses both grippers to cooperatively manipulate a flexible sheet, lift the herbs, move them above a target container, and pour them into it. This task requires stable bimanual coordination throughout the trajectory, since the support sheet deforms during lifting and tilting. It is also sensitive to subtle contact changes during the final pouring stage, where the robot must release the remaining herbs by flattening the paper while maintaining enough support to avoid spillage or tearing. We therefore report both whole-task success and stage-wise success. The key intermediate stages include successful grasping of the paper, stable transfer to the target region, and successful pouring into the container.

\textbf{Cable mounting.}
In this task, the robot lifts a flexible cable, aligns it with a target clip, and presses it into place. This task presents two key challenges. First, the cable may shift or slip during gripper closure, which makes the pickup stage sensitive to contact quality. Second, successful seating is not always easy to determine from vision alone, especially when the cable and clip have similar appearance or low visual contrast. The evaluation therefore includes both whole-task success and stage-wise metrics. We specifically report whether the robot successfully picks up the cable and whether it seats the cable fully into the clip. These intermediate metrics help distinguish failures caused by unstable grasping from those caused by contact-rich insertion.

\textbf{Binder clip removal.}
This task requires the robot to grasp a spring-loaded binder clip attached to a folder, detach it, open a drawer, place the clip inside, and then close the drawer. Compared with the other tasks, this task involves both a contact-rich release action and a longer sequential manipulation chain. The initial detachment stage is particularly sensitive to grasp stability, since the resistance of the spring-loaded clip changes during removal. Tactile feedback is helpful for maintaining a secure grasp and identifying whether the clip has been detached successfully. We therefore evaluate both the initial clip-detachment success and the final whole-task success after the drawer operation is completed. In the generalization setting, where the drawer appearance is changed, we additionally report the success rate of the drawer-opening stage to directly assess how well the policy transfers to this altered condition.

\begin{figure}[!t]
      \centering
      \includegraphics[width=\linewidth]{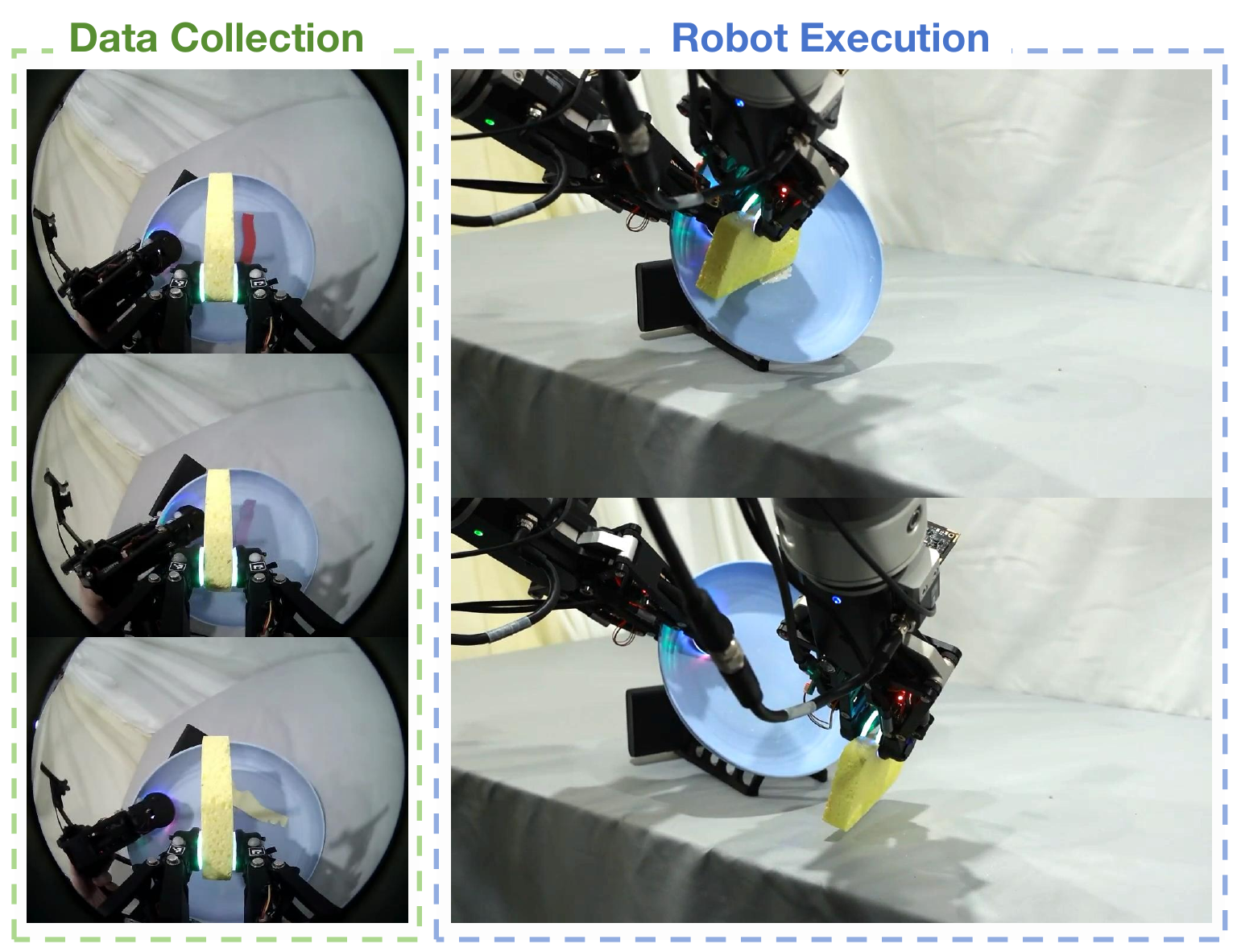}
      \caption{\textbf{Dish washing from data collection to robot execution.} Demonstrations collected with proxy materials are successfully transferred to real cleaning scenarios, enabling stable contact and effective wiping on real stains.
 }
      \label{fig:dishwashingtask}
   \end{figure}

\textbf{Dish washing.}
In this task, the robot grasps a dish and a sponge, places the sponge onto the stained region, and performs wiping until the stain is removed. This task emphasizes coordinated dual-object manipulation under sustained contact. Success depends not only on grasping the two objects, but also on establishing and maintaining effective contact between the sponge and the dish surface during the wiping motion. Since stable contact and friction are critical in this task, tactile observations provide useful cues beyond visual appearance alone. To avoid material waste during data collection, demonstrations are collected using strip-based props with different colors. During evaluation, however, the real cleaning behavior is tested using white jam on the dish surface, as shown in~\cref{fig:dishwashingtask}.

\textbf{Success criteria.}
A trial is counted as a successful whole-task execution only if the robot completes the entire task objective without human intervention. In addition to whole-task success, we also report stage-wise success for key intermediate phases that are most sensitive to tactile feedback and failure recovery. For herbal transfer, these phases include grasping and pouring. For cable mounting, they include cable pickup and full seating. For binder clip removal, they include clip detachment and, in the generalization setting, drawer opening. These finer-grained metrics help identify whether a method improves initial contact establishment and subsequent contact-rich manipulation.

\subsection{Training and Implementation Details}
\textcolor{gray}{\textit{| Supplement to \cref{sec:method-policy} in the Main paper.}}

\textbf{Policy architecture.}
Our downstream policy follows an ACT-style transformer architecture. Visual and tactile observations are encoded separately using ResNet-18 backbones. Their features are projected into a shared latent space and fused before transformer-based action prediction. The transformer uses a 4-layer encoder and a 7-layer decoder, with a hidden dimension of 512 and a feedforward dimension of 3200. The policy outputs a 16-dimensional action vector corresponding to the dual-arm joint commands and gripper actions.

\textbf{Training objectives and optimization.}
Training proceeds in three stages: tactile representation pretraining, task-specific bimanual imitation learning, and recovery-based refinement. In the pretraining stage, the tactile encoder is initialized using the contrastive objective described in the main paper. The downstream ACT policy is then trained with supervised action prediction on task-specific bimanual demonstrations. Since the four tasks differ in temporal horizon and motion complexity, the action chunk size is selected separately for each task. For all downstream experiments, we use a learning rate of \(1\times10^{-5}\) and a KL weight of 10. Training is performed on trajectory sequences rather than independently sampled frames.

\textbf{Task-specific training data.}
The task-specific training data are summarized in Table~\ref{tab:training_data}. The number of bimanual demonstrations varies across tasks according to task complexity. 
For tasks with recovery-based refinement, we further collect recovery trajectories from representative policy-induced failure cases. In herbal transfer, these failures mainly arise from unsuccessful grasping and pouring. In cable mounting, recovery data cover several common failure modes, including failing to grasp the cable, dropping it during transport and insertion, and misaligned placement during insertion. In binder clip removal, recovery cases include both grippers failing to establish a grasp simultaneously, as well as unsuccessful clip grasping. In dish washing, recovery data are collected for failures such as missing the sponge, missing both dish, and requiring repeated wiping. These recovery trajectories enrich the dataset with corrective behaviors near realistic failure states and support more robust policy refinement.
\begin{table}[t]
  \centering
  \caption{
    \textbf{Task-specific training data.}
    The table summarizes the number of bimanual demonstrations and recovery trajectories used for downstream policy training.
  }
  \label{tab:training_data}
  \setlength{\tabcolsep}{6pt}
  \begin{tabular}{lcc}
    \toprule
    \textbf{Task} & \makecell{Bimanual\\Demonstrations} & \makecell{Recovery\\Trajectories} \\
    \midrule
    Herbal Transfer      & 94  & 10 \\
    Cable Mounting       & 221 & 21 \\
    Binder Clip Removal  & 107 & 10 \\
    Dish Washing         & 98  & 10 \\
    \bottomrule
  \end{tabular}
\end{table}
\newpage

\vfill

\end{document}